%% file: example.tex
\documentclass[11pt]{article}

\PassOptionsToPackage{table}{xcolor}
\usepackage[preprint]{acl}

\usepackage{times}
\usepackage{latexsym}
\usepackage[T1]{fontenc}
\usepackage[utf8]{inputenc}
\usepackage{inconsolata}
\usepackage{microtype}
\usepackage{graphicx}
\usepackage{subcaption}
\usepackage{booktabs}
\definecolor{lightblue}{RGB}{219,238,242}

\usepackage{amsmath}
\usepackage{amssymb}
\usepackage{mathtools}
\usepackage{amsthm}
\usepackage{multirow}
\usepackage{makecell}
\usepackage[most]{tcolorbox}
\usepackage{pifont}
\usepackage[textsize=tiny]{todonotes}
\usepackage[capitalize,noabbrev]{cleveref}

\newcommand{\cmark}{\ding{51}}
\newcommand{\xmark}{\ding{55}}

\theoremstyle{plain}

\theoremstyle{definition}

\theoremstyle{remark}

\title{ViCA: Efficient Multimodal LLMs with Vision-Only Cross-Attention}

\author{
 \textbf{Wenjie Liu\textsuperscript{1}\footnotemark[1]},
 \textbf{Hao Wu\textsuperscript{1}\thanks{Equal contribution}},
 \textbf{Xin Qiu\textsuperscript{1}},
 \textbf{Xudong Wang\textsuperscript{1}},
\\
 \textbf{Yingqi Fan\textsuperscript{1}},
 \textbf{Yihan Zhang\textsuperscript{1}},
 \textbf{Anhao Zhao\textsuperscript{1}},
 \textbf{Yunpu Ma\textsuperscript{2}},
 \textbf{Xiaoyu Shen\textsuperscript{1}\thanks{Corresponding author}}
\\
 \textsuperscript{1}Ningbo Institute of Digital Twin, Eastern Institute of Technology, Ningbo \\
 \textsuperscript{2}Munich Center for Machine Learning, LMU Munich
\\
   \href{mailto:email@domain}{xyshen@eitech.edu.cn}      
}

\begin{document}
\maketitle

\begin{abstract}
Modern multimodal large language models (MLLMs) adopt a unified self-attention design that processes visual and textual tokens at every Transformer layer, incurring substantial computational overhead. In this work, we revisit the necessity of such dense visual processing and show that projected visual embeddings are already well-aligned with the language space, while effective vision-language interaction occurs in only a small subset of layers. Based on these insights, we propose \textbf{ViCA} (Vision-only Cross-Attention), a minimal MLLM architecture in which visual tokens bypass all self-attention and feed-forward layers, interacting with text solely through sparse cross-attention at selected layers. Extensive evaluations across three MLLM backbones, nine multimodal benchmarks, and 26 pruning-based baselines show that ViCA preserves \textbf{98}\% of baseline accuracy while reducing visual-side computation to \textbf{4}\%, consistently achieving superior performance-efficiency trade-offs. Moreover, ViCA provides a regular, hardware-friendly inference pipeline that yields $>$\textbf{3.5}$\times$ speedup in single-batch inference and $>$\textbf{10}$\times$ speedup in multi-batch inference, reducing visual grounding to \emph{near-zero} overhead compared with text-only LLMs. It is also orthogonal to token pruning methods and can be seamlessly combined for further efficiency gains. The code is released at \url{https://github.com/EIT-NLP/ViCA}.
\end{abstract}

\input{Sec/1_introduction}
\input{Sec/6_related_work}
\input{Sec/2_preliminary}

\input{Sec/4_method}

\input{Sec/5_experiment}

\input{Sec/7_conclusion}
\input{Sec/8_limitation}


\bibliography{example_paper}

\newpage
\appendix
\input{Sec/x_supp}

\end{document}

%% file: Sec/1_introduction.tex
\section{Introduction}
\label{sec:intro}

Multimodal large language models (MLLMs) extend large language models (LLMs) to visual contexts, enabling models to understand and respond to visual inputs~\citep{gpt4v, li2023blip, dai2023instructblip, lin2024vila}. Early MLLMs often injected visual representations into language models through \emph{cross-attention}~\citep{alayrac2022flamingo, li2022blip}, whereas recent MLLMs have largely adopted a unified \emph{self-attention} design: visual features are projected into the language embedding space, concatenated with text tokens, and processed by every Transformer layer~\citep{liu2023improvedllava, liu2024llavanext,2024internvl2,wang2025internvl3,bai2025qwen2}. While conceptually simple and highly extensible, this design is computationally inefficient: visual tokens are repeatedly updated by self-attention and feed-forward networks (FFNs) at every layer, resulting in substantial vision-side overhead~\citep{Wu_2026, shao2026a, yang2024visionzip}.

To mitigate this cost, recent work has focused on pruning-based acceleration, including token-level methods that remove less important visual tokens~\citep{zhang2024sparsevlm, shang2025llava-prumerge} and operation-level methods that skip vision-related operations~\citep{lin2025VTW, wu2026hidrop}. However, most such approaches still operate within the unified self-attention architecture and face two practical limitations. First, many rely on input-dependent importance estimation, which can be sensitive to visual redundancy, positional effects, and cross-layer attention variability, making pruning decisions unstable~\citep{kim2025training, wen2025stop}. Second, pruning incurs extra system overhead (e.g., importance scoring, selection and reordering), leading to nonlinear latency behavior and causing practical speedups to fall short of theoretical FLOPs reductions~\citep{wen-etal-2025-token}\footnote{These overheads are dominated by non-arithmetic GPU costs (e.g., kernel launches, gather/scatter, buffering and synchronization)~\citep{fernandez-etal-2023-framework}. Even small token-count changes can disrupt kernel grids and FlashAttention tiling, often worsened by GPU tail effect~\citep{eliopoulos2025pruning}.}.

Meanwhile, prior studies have revealed substantial redundancy in vision-side computation: visual attention is largely local~\citep{zhang2024treat, li2025redundancylens}, only a small fraction of FFN channels significantly affects performance~\citep{yuan2025shortv, fan2025visipruner}, and freezing visual updates in certain layers incurs minimal accuracy loss even without training~\citep{yuan2025shortv}. These observations raise a more structural question: \emph{are repeated visual-token write operations necessary in unified self-attention MLLMs, or can they be removed through architectural operation pruning?}

In this work, we empirically test this hypothesis by freezing visual tokens immediately after projection, completely removing their self-attention and FFN updates. Surprisingly, this aggressive simplification results in only a marginal performance drop, indicating that \emph{projected visual embeddings are already well aligned with the LLM's semantic space}. Thus, repeated self-attention-based refinement of visual tokens provides little additional benefit. To further locate the necessary cross-modal read paths, we analyze layer-wise cosine similarity of text representations before and after text-to-vision cross-attention, finding that effective multimodal interaction is concentrated in only a few layers.

Based on these findings, we propose \textbf{ViCA} (Vision-only Cross-Attention), a minimal and efficient MLLM architecture that reinterprets visual tokens as read-only memory for the LLM. Inspired by early cross-attention-based MLLMs but derived from redundancy diagnostics in modern self-attention MLLMs, ViCA removes visual-token self-attention and FFN updates, preserving vision--language interaction only through sparse text-to-vision cross-attention at a small set of key layers. We validate ViCA across 3 representative backbones and 9 widely used multimodal benchmarks. The results show that ViCA preserves 98\% of baseline accuracy, while reducing visual-side computation to 4\% of the original architecture. Compared with \emph{26} existing pruning methods covering both token-level and operation-level strategies, ViCA consistently achieves superior performance--efficiency trade-offs, demonstrating that \emph{architecturally eliminating redundant visual computation is more effective than dynamically pruning it within a self-attention-based framework}.

Importantly, ViCA introduces an asymmetric attention structure: queries originate exclusively from text tokens, while keys and values come from both text and visual tokens. This structure is naturally compatible with FlashAttention\footnote{FlashAttention natively supports short-query--long-key/value causal attention by aligning the causal mask to the bottom-right corner of the attention matrix~\citep{dao2024flashattention}. Thus, ViCA can use standard FlashAttention kernels without materializing full attention maps or suffering from pruning-induced kernel/tiling degradation.} and avoids the dynamic selection and reordering overhead of token pruning, bringing multimodal prefill latency close to the text-only setting. Moreover, ViCA is orthogonal to token-dropping methods and can be combined with them for further efficiency gains. In summary, our main contributions are as follows:

\noindent (1) \textbf{Architectural necessity analysis.}  We show that visual-token updates via self-attention and FFNs are largely unnecessary in self-attention MLLMs, while effective cross-modal reads are concentrated in only a few layers.

\noindent (2) \textbf{Minimal and efficient architecture.} Based on these insights, we propose ViCA (Vision-only Cross-Attention), which eliminates redundant computation and enables a fully regular, hardware-friendly inference pipeline.

\noindent (3) \textbf{Extensive empirical validation.} We evaluate ViCA across three backbones, nine multimodal benchmarks, and 26 pruning methods. ViCA achieves the best accuracy--efficiency trade-off, preserving $\sim$98\% of baseline accuracy while reducing visual-side computation to 4\%, with over $3.5\times$ single-batch and $10\times$ multi-batch prefill speedups.

%% file: Sec/6_related_work.tex
\section{Related Work}
\label{sec:rel}

Visual token pruning reduces the number of visual tokens processed by the LLM, typically by ranking and removing low-saliency tokens. Existing methods estimate token importance from text-to-vision attention~\cite{chen2024image, xing2024pyramiddrop, zhang2024sparsevlm}, vision-to-vision attention~\cite{ye2025fit, han2024filter, ye2025atp, yin2025lifting}, or model-guided signals~\cite{zhao2025stitch, huang2024dynamic, sun2025lvpruning, shao2025growing}. To mitigate information loss under aggressive pruning, some methods merge discarded tokens into compact representatives~\cite{dhouib2025pact, tan2025tokencarve, zhang2024sparsevlm}.

While visual token pruning reduces computation at the sequence level by shortening visual sequence, operation pruning reduces computation at the operator level by sparsifying internal attention and FFN computations~\cite{zhao2025skipgpt, han2025informed, qiu2025few, ding2026llms}. Existing methods exploit visual computation redundancy from depth-wise and width-wise perspectives. Depth-wise methods reduce visual computation across layers. ShortV~\citep{yuan2025shortv} diagnoses redundant visual-token updates via layer-wise freezing. Skip-Vision~\cite{zeng2025skip} and DOP~\cite{liu2025fine} skip FFN or both attention and FFN operations for visual tokens, while HiDrop~\citep{wu2026hidrop} allows visual tokens to enter late and exit early. Width-wise methods such as YOPO~\citep{zhang2024treat} and RedundancyLens~\citep{li2025redundancylens} sparsify within-layer visual computation by restricting vision-to-vision attention and pruning FFN dimensions. 
Different from these heuristic pruning designs, our work turns attention- and FFN-level redundancy diagnostics into a compact necessary visual-update architecture and validates each retained path through experiments.

%% file: Sec/2_preliminary.tex
\section{Information-Flow View: Self-Attention vs. Cross-Attention Interaction} 
\label{sec:pre}

\begin{figure}[t]
    \centering
    \includegraphics[width=.95\linewidth]{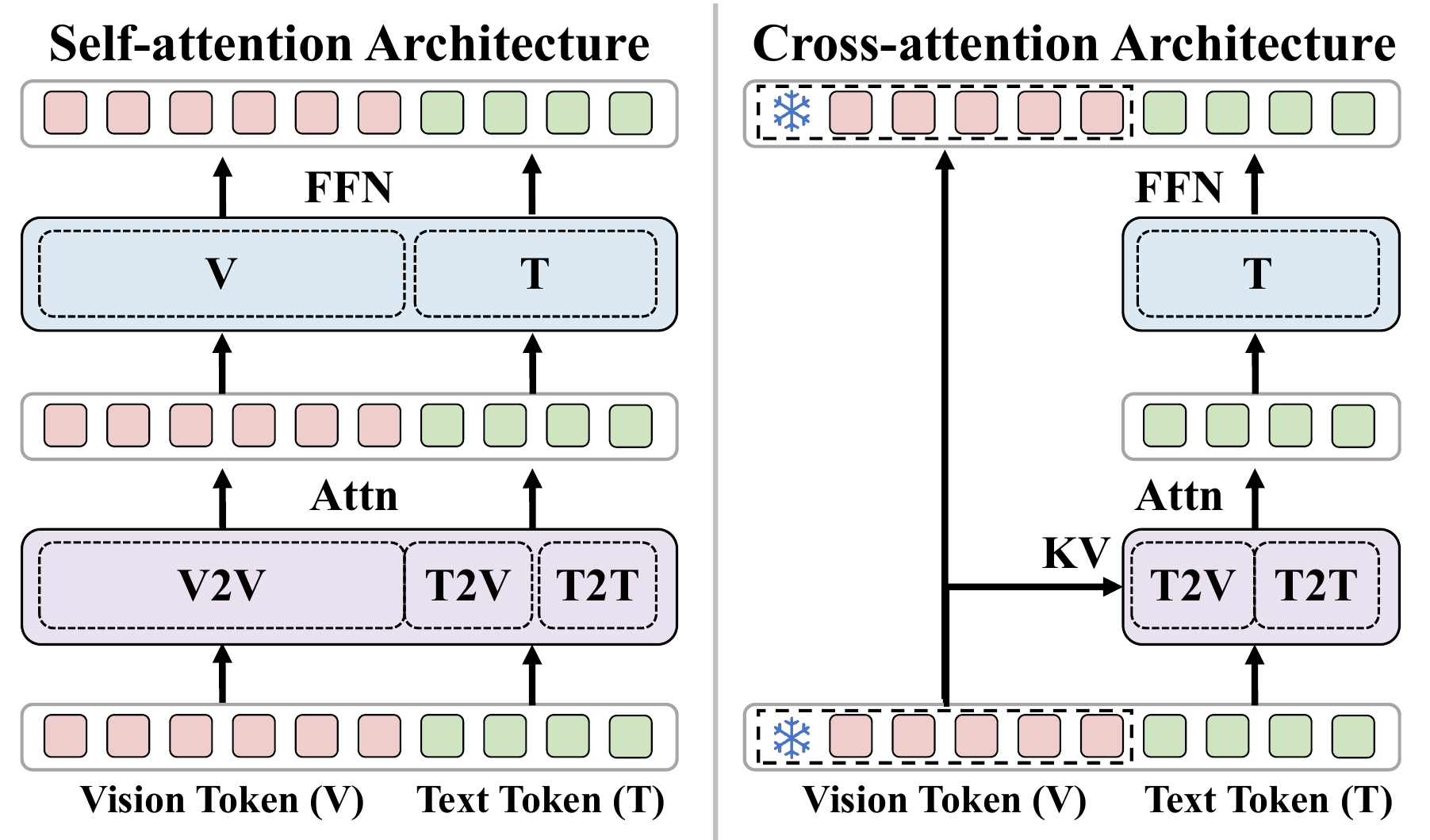}
    \caption{\small Comparison between self-attention and cross-attention architectures from an information-flow perspective. Attention and FFN operations for visual tokens dominate the computation in MLLM. V2V, T2V, and T2T denote vision-to-vision, text-to-vision, and text-to-text attention, respectively.}
    \label{fig:self-vs-cross}
    \vspace{-10pt}
\end{figure}

Building on prior efficient MLLM studies~\cite{zhang2025cross, yin2025lifting, yang2025vflowopt, li2025redundancylens, tong2025flowcut}, we present an information-flow view to precisely define what is being pruned. We distinguish \textbf{self-attention} and \textbf{cross-attention} interaction patterns by how visual tokens participate in attention and FFN computation. Let $H^{(l)}=[V^{(l)};T^{(l)}]$ be the concatenation of visual and text tokens at layer $l$. A self-attention layer can be written in a blockwise form as follows:
\begin{equation*}
\resizebox{0.9\linewidth}{!}{$
\begin{gathered}
\tilde{H}^{(l)}
= \begin{bmatrix} \tilde{V}^{(l)} \\ \tilde{T}^{(l)} \end{bmatrix}
= H^{(l)} +
\begin{bmatrix}
A_{VV} & A_{VT} \\
A_{TV} & A_{TT}
\end{bmatrix}
\begin{bmatrix}
V^{(l)} W_V \\
T^{(l)} W_V
\end{bmatrix} \\
H^{(l+1)}
= \tilde{H}^{(l)} +
\begin{bmatrix}
\mathrm{FFN}_V(\tilde{V}^{(l)}) \\
\mathrm{FFN}_T(\tilde{T}^{(l)})
\end{bmatrix}
\end{gathered}
$}
\end{equation*}
where $A=\mathrm{softmax}(QK^\top/\sqrt{d})$ is induced by masked self-attention over the unified sequence (we omit multi-head/output projections for brevity). All tokens share the same value projection $W_V$, following standard self-attention; the block form only reflects token-level information flow. In the common decoder-only setting with a causal mask and vision tokens placed as a prefix, the mask enforces $A_{VT}=0$, leaving three active channels: V2V ($A_{VV}$), T2V ($A_{TV}$), and T2T ($A_{TT}$). As illustrated in Figure~\ref{fig:self-vs-cross}, \textbf{self-attention} performs joint \emph{read--write} over $[V;T]$: visual tokens are \emph{updated} via V2V attention and the visual FFN branch $\mathrm{FFN}_V(\cdot)$, while text tokens \emph{read} vision through $A_{TV}$ and interact within text via $A_{TT}$ and $\mathrm{FFN}_T(\cdot)$.

In contrast, \textbf{cross-attention} treats visual tokens as \emph{read-only memory}: text queries attend to visual keys/values, while visual tokens are \emph{frozen} and not updated by attention or FFN, i.e., enforcing $\Delta V^{(l)}=0$ by bypassing all visual write operations. This ``freeze-vision'' intuition is supported by prior evidence that \emph{visual-token writes have heavy redundancy}~\cite{yuan2025shortv}. Different from such training-free analyses, our approach is training-based and leverages this redundancy to redesign interaction paths\footnote{In Appendix~\ref{app:visual-update-ratio} and Table~\ref{tab:visual-update-ratio} detail the FLOPs calculation, showing that visual-write operations dominate vision-side FLOPs (often over 80\%), highlighting the efficiency advantage of such read-only architecture.}. This read-only design is conceptually related to early cross-attention-based MLLMs, such as Flamingo~\citep{alayrac2022flamingo} and BLIP~\cite{li2022blip}, where vision features are computed once and exposed to the language model through cross-attention as KV memory. While such early MLLMs avoid repeated visual updates, they typically rely on modality-specific cross-attention or two-stream architectures, which limit deep fusion and require careful architectural customization. Consequently, recent MLLMs have largely converged to self-attention-based designs that unify vision and language tokens into a single sequence, enabling stronger cross-modal interaction with greater architectural simplicity. Our analysis revisits this trade-off and motivates architectures that retain unified modeling while avoiding redundant visual-token writes.

%% file: Sec/4_method.tex
\section{Methodology}
\label{sec:method}


\subsection{Where Visual Information Matters}

Following the information-flow view in Section~\ref{sec:pre}, we empirically study how visual-token \emph{writes} and cross-modal \emph{reads} affect model outputs.

\paragraph{Impact Measurement of Visual Updates.}
To operationalize the information-flow perspective, we use two metrics to capture representation change and output impact. Representation change is measured by the cosine similarity between token embeddings before and after a module:
\begin{equation*}
    \scalebox{0.95}{$
        \mathrm{Cos}(h, h') = \frac{h^\top h'}{\|h\| \, \|h'\|}
    $}
\end{equation*}
where $h$ and $h'$ denote token embeddings before and after the update.
Output impact is measured by the KL divergence between the original output distribution $p(y)$ and the distribution obtained after disabling the update:
\begin{equation*}
    \scalebox{0.95}{$
        \mathrm{KL}(p \,\|\, p') = \sum_y p(y) \log \frac{p(y)}{p'(y)}
    $}
\end{equation*}
Intuitively, cosine similarity quantifies how much representations are modified, while KL divergence indicates whether such modifications propagate to the final output distribution.

\begin{figure}[t]
\centering
\includegraphics[width=.95\linewidth]{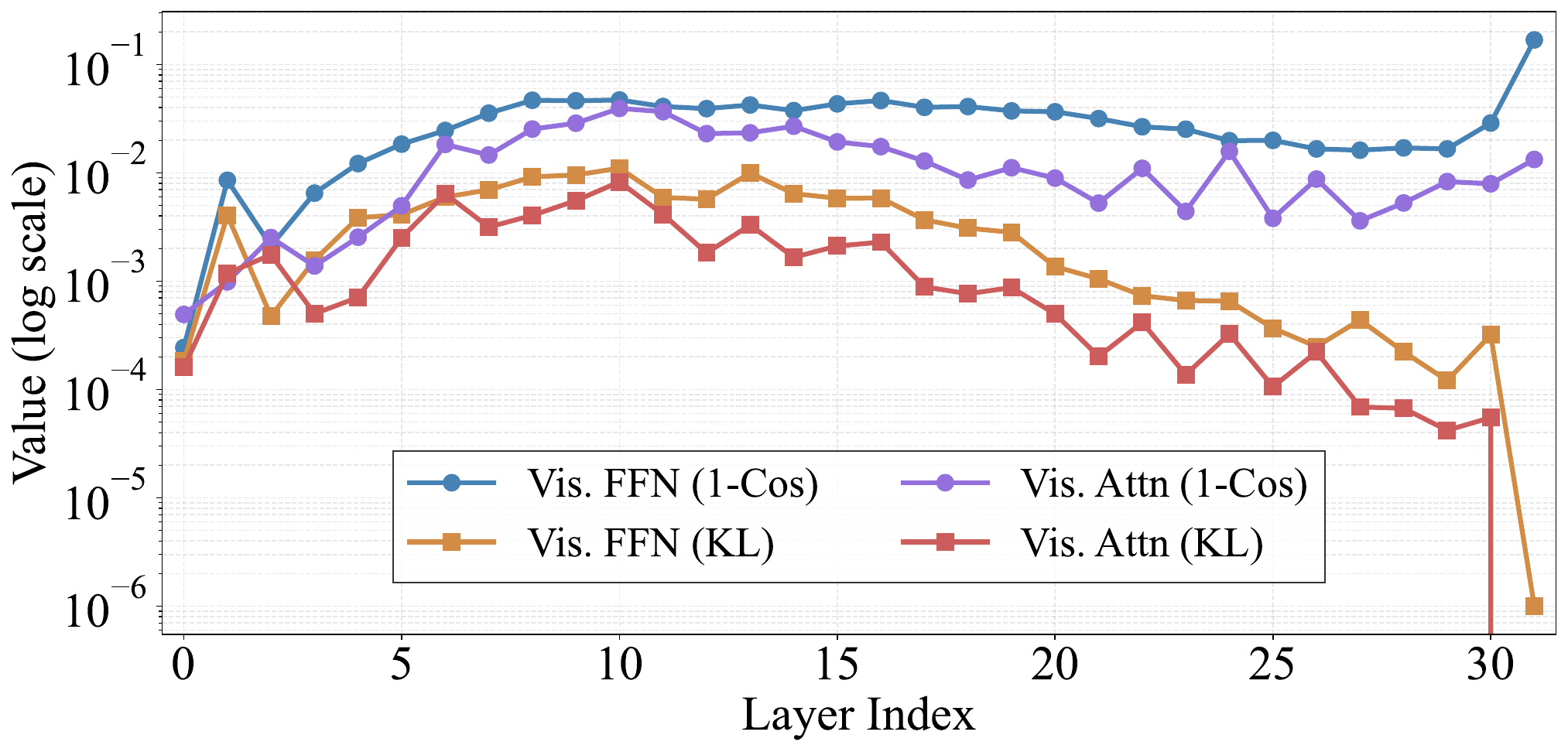}
\vspace{-2pt}
\caption{\small Layer-wise diagnostics of visual token updates in LLaVA-1.5-7B on TextVQA, measured by output impact (KL) and representation change (1-Cos).}
\label{fig:kl_cossim_vis-update}
\vspace{-5pt}
\end{figure}

\begin{figure}[t]
\centering
\includegraphics[width=.95\linewidth]{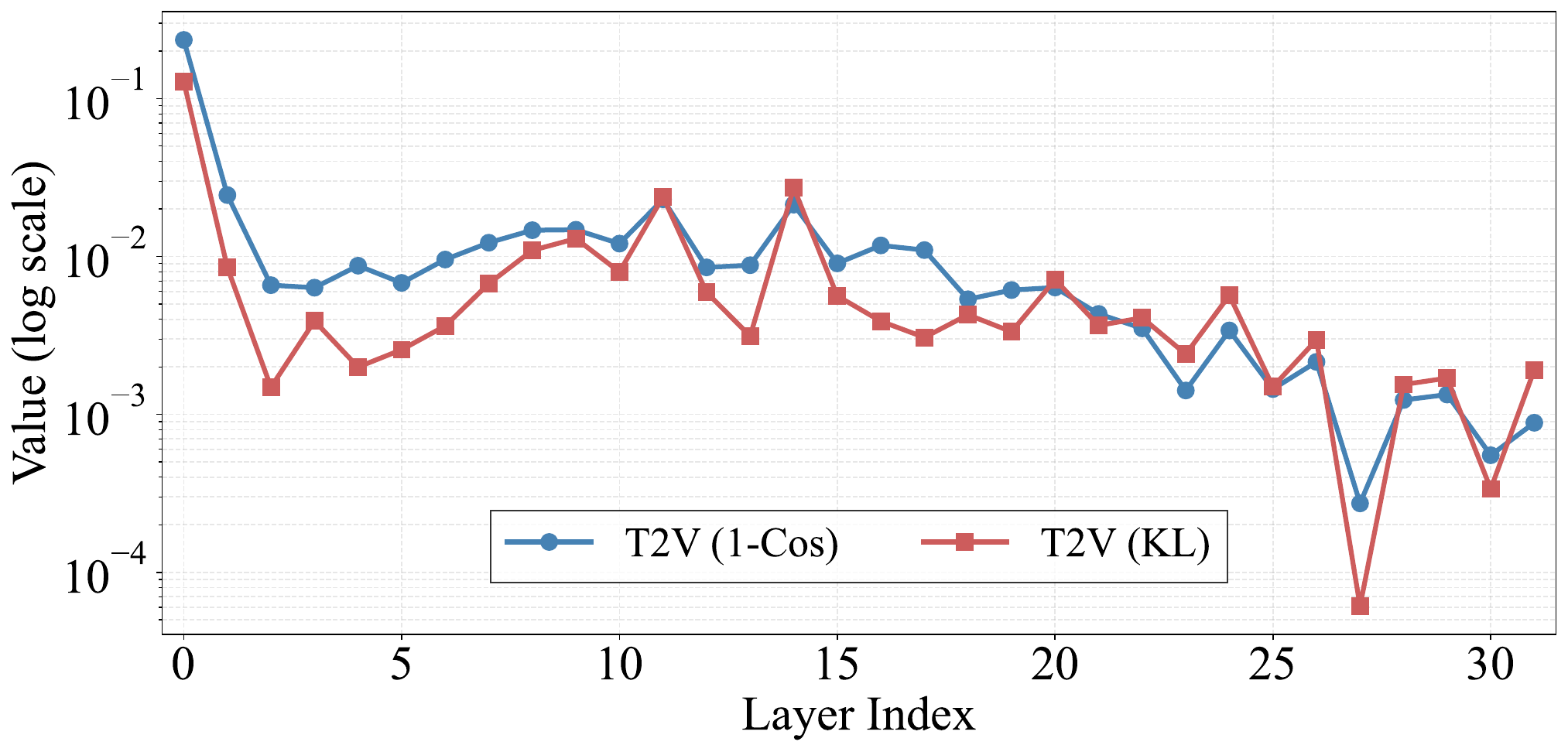}
\vspace{-2pt}
\caption{\small Layer-wise diagnostics of text--to-vision (T2V) cross-attention in LLaVA-1.5-7B on TextVQA, measured by output impact (KL) and representation change (1-Cos).}
\label{fig:t2v-kl-cos}
\vspace{-5pt}
\end{figure}

\begin{table}[t]
\centering
\resizebox{\columnwidth}{!}{
\begin{tabular}{ccc|cccc|c}
\toprule
\textbf{Setup} & \textbf{\makecell{Retained\\Layers}} & \textbf{\makecell{Layer\\Indices}} & \textbf{MME$^{P}$} & \textbf{GQA} & \textbf{VQA\textsuperscript{T}} & \textbf{POPE} & \textbf{Avg. (\%)} \\
\midrule
\addlinespace[0.4ex] \rowcolor{gray!20}
\multicolumn{8}{c}{\textbf{Baseline}} \\
\midrule
- & Full & 0--31 & 1506.5 & 61.9 & 58.2 & 86.8 & 100.0\\
\midrule
\addlinespace[0.4ex] \rowcolor{gray!20}
\multicolumn{8}{c}{\textbf{Vision-Update Attention}} \\
\midrule
\multirow{4}{*}{TF} & Essential     & 6--11             & 1454.5 & 61.0 & 56.1 & 86.7 & 98.3\\ 
                    & Over-approx.  & 6--17             & 1469.7 & 61.6 & 56.6 & 86.0 & 98.8\\ 
                    & Non-essential & 0-5,12-31         & 1361.2 & 59.2 & 55.4 & 83.9 & 95.3\\   
                    & Frozen        & -                 & 1298.2 & 54.5 & 51.0 & 79.8 & 91.7\\ 
\midrule
T & Frozen & - & 1501.1 & 62.1 & 57.1 & 87.1 & 100.2 \\
\midrule
\addlinespace[0.4ex] \rowcolor{gray!20}
\multicolumn{8}{c}{\textbf{Visual-Update FFN}} \\
\midrule
\multirow{3}{*}{TF} & Essential     & 5--11             & 1463.2 & 58.9 & 54.2 & 85.2 & 96.8\\     
                    & Non-essential & 0-4,12-31         & 988.6  & 49.8 & 48.9 & 65.4 & 81.8\\  
                    & Frozen        & -                 & 876.8  & 39.6 & 40.9 & 61.1 & 71.2\\      
\midrule
\multirow{2}{*}{T}  & Essential     & 5--11             & 1475.7 & 63.3 & 58.2 & 87.3 & 100.5\\     
                    & Frozen        & -                 & 1472.4 & 62.0 & 56.9 & 87.3 & 99.2\\     
\midrule
\addlinespace[0.4ex] \rowcolor{gray!20}
\multicolumn{8}{c}{\textbf{Text-to-Vision Cross-Attention}} \\
\midrule
\multirow{3}{*}{TF} & Essential     & 0--1,7--11,14     & 1393.5 & 54.4 & 51.3 & 86.1 & 93.5 \\ 
                    & Non-essential & 2-6,12-13,15-31   & 758.1  & 44.8 & 45.4 & 70.4 & 63.0 \\  
                    & Frozen        & -                 & 572.1  & 29.9 & 29.9 & 50.3 & 46.2 \\ 
\midrule
T                    & Essential    & 0--1,7--11,14     & 1476.6 & 62.7 & 56.4 & 87.1 & 99.5\\
\bottomrule
\end{tabular}
}
\caption{\small Redundancy diagnosis in LLaVA-1.5-7B (TF: training-free; T: training-based). Freezing all vision-update attention/FFN operations and retaining T2V attention only in essential layers cause negligible performance change after training. See Appendix Table~\ref{tab:llava7b_trainfree} for details.}
\label{tab:llava-key-layers}
\vspace{-10pt}
\end{table}

\paragraph{Redundancy in Visual-Token Write.}
We rank Transformer layers using the two metrics defined above. As shown in Figure~\ref{fig:kl_cossim_vis-update}, cosine similarity detects broad representation changes in early-to-mid layers (attention: 6--17; FFN: 5--31), whereas KL diagnostics identify a much narrower output-critical subset (attention: 6--11; FFN: 5--11). Beyond these layers, visual representations still change but have negligible impact on predictions. We therefore evaluate three regimes: \emph{full}, \emph{essential}, and \emph{none}, under both training-free (TF) and training-based (T) protocols, with results in Table~\ref{tab:llava-key-layers}. In the training-free setting, retaining only essential layers largely preserves performance; with training-based adaptation, performance is nearly restored even when all visual-token updates are frozen. These results suggest that \emph{many visual-token updates, especially in deeper layers, are functionally redundant, and that fine-tuning can compensate for their removal}.

\paragraph{Redundancy in Cross-Modal Read.}
Prior studies suggest that visual information enters language representations mainly through text--vision (T2V) cross-attention~\citep{zhang2025crossmodal_information_flow}. Consistently, Figure~\ref{fig:t2v-kl-cos} shows that effective cross-modal injection is concentrated in a few early-to-mid layers, peaking around layer 10 in LLaVA-1.5-7B. After these layers, later visual-token updates are rarely read by text tokens, explaining their limited effect on final outputs despite changing representations. Table~\ref{tab:llava-key-layers} further shows that restricting T2V cross-attention and visual updates to these essential layers largely preserves performance. These results suggest that \emph{once visual information is injected through a few cross-attention layers, subsequent visual-token updates become largely redundant}.

\subsection{ViCA: Vision-Only Cross-Attention}

\begin{figure}[t]
\centering
\includegraphics[width=\linewidth]{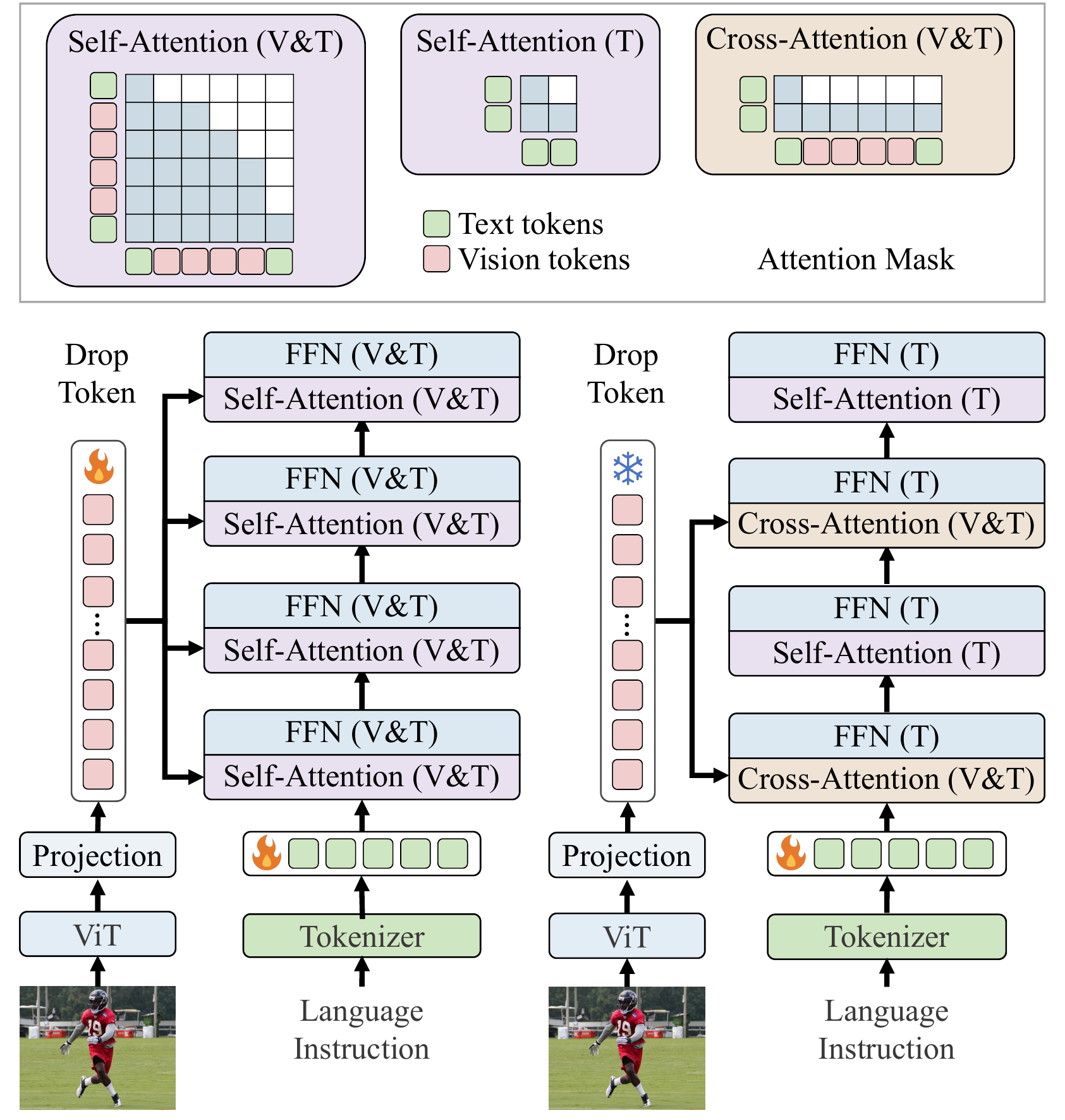}
\caption{\small Common token dropping vs. our minimal architecture (ViCA). Left: Token dropping removes some visual tokens, but the remaining ones still undergo full self-attention and FFN updates across layers, incurring substantial visual computation. Right: ViCA removes visual update paths in attention and FFN. Visual tokens act only as KVs in a few cross-attention layers, while all other operations run on text tokens, reducing visual computation.}
\label{fig:vica}
\vspace{-10pt}
\end{figure}

Motivated by the above diagnostics, we introduce \textbf{ViCA}, a minimal architectural modification. Figure~\ref{fig:vica} contrasts ViCA with token-dropping approaches and illustrates its design:

(1) \textbf{Removing visual token write operations}: Visual tokens no longer participate in self-attention or FFN updates at any transformer layer. After the initial projection, visual token representations remain fixed and are not written by either attention or FFN modules. Importantly, this differs from freezing the vision encoder: ViCA operates entirely within the language model and does not prevent the vision encoder itself from being finetuned.

(2) \textbf{Sparse cross-modal read operations}: To preserve effective vision-to-language read path, text--to-vision cross-attention is retained only at a small set of diagnostically identified key layers. In these layers, visual tokens serve as static keys and values that text tokens can attend to, while all remaining layers operate purely on text tokens.

Together, these designs yield a streamlined computation graph in which \emph{visual-token writes are entirely removed and cross-modal read is confined to a few critical layers}. This restructuring eliminates the dominant sources of redundant visual computation during pre-filling, substantially reducing vision-side cost while preserving the cross-modal information flow necessary for performance.

\subsection{Practical Acceleration and Compatibility}
\paragraph{Comparison with Token Pruning.}
Most token-pruning methods estimate acceleration using token counts or FLOPs, yet these metrics fail to capture dominant GPU overheads, including kernel launches, data movement, and synchronization~\cite{fernandez-etal-2023-framework}, as well as the runtime cost of pruning itself, such as importance scoring and token selection~\cite{wen-etal-2025-token}. Moreover, dynamically changing token counts induce frequent reconfiguration of attention kernels and FlashAttention tiling, leading to highly nonlinear latency behavior that often decouples practical speedup from theoretical FLOPs reduction~\cite{eliopoulos2025pruning, nvidia_gpu_performance_guide}.
In contrast, ViCA targets the computation \emph{architecture} rather than the sequence length. By keeping token counts and attention shapes fixed across layers while removing visual-token write paths (i.e., excluding visual tokens from attention queries and FFNs), ViCA avoids dynamic shapes, kernel reconfiguration, and unnecessary data movement. As a result, the reduced computation maps more directly and reliably to real-world inference speedups.

\paragraph{Compatibility with FlashAttention.}
In ViCA, some layers perform text-only attention. In other layers, queries come from text tokens, and KVs include both text and visual tokens. Since FlashAttention v2.1+ supports this ($\text{q}_{len} < \text{kv}_{len}$) pattern under causal masking by aligning the mask to the bottom-right corner. By simplifying the vision computation graph without modifying token sequence lengths or attention tensor shapes, ViCA preserves full compatibility with efficient attention kernels and existing inference optimizations. (see Appendix~\ref{app:flash-atten---imp} for details).

\paragraph{Orthogonality to Token Pruning.}
ViCA prunes redundant visual computation paths instead of tokens, making it orthogonal to existing token-dropping approaches. These methods can be combined at inference time to further reduce visual computation with minimal performance impact. Unlike most token pruning methods, which assume visual importance decreases with depth~\cite{xing2025pyramid_drop}, ViCA restricts visual tokens to cross-attention in fewer layers, avoiding unnecessary retention of tokens in layers where they provide little benefit. This makes ViCA a more effective and interpretable foundation for token pruning.

%% file: Sec/5_experiment.tex
\section{Experiments}
\label{sec:Experiments}
\begin{table*}[t]
\centering
\resizebox{0.97\textwidth}{!}{
    \begin{tabular}{c|cc|cc|ccccccccc|c}
    \toprule
    \multirow{2}{*}{\textbf{Method}} &
    \multicolumn{2}{c|}{\textbf{Sparsity}} &
    \multicolumn{2}{c|}{\textbf{Vision-side}} &
    \multirow{2}{*}{\textbf{MME\textsuperscript{P}}} & 
    \multirow{2}{*}{\textbf{MMB}} &  
    \multirow{2}{*}{\textbf{MMB\textsuperscript{CN}}} &  
    \multirow{2}{*}{\textbf{GQA}} &  
    \multirow{2}{*}{\textbf{VQA\textsuperscript{v2}}} & 
    \multirow{2}{*}{\textbf{SQA\textsuperscript{I}}} &  
    \multirow{2}{*}{\textbf{VQA\textsuperscript{T}}} & 
    \multirow{2}{*}{\textbf{POPE}} &  
    \multirow{2}{*}{\textbf{SEED\textsuperscript{I}}} &
    \multirow{2}{*}{\textbf{Rel. Avg.}} \\ 
    & \textbf{Op.} & \textbf{Tok.} & \textbf{Token} & \textbf{TFLOPs (Rel.)} & & & & & & & & & \\
    \midrule
    LLaVA-1.5-7B                & -      & -      & 576 & 7.65 (100.0\%) & 1506.5 & 64.7 & 58.1 & 61.9 & 78.5 & 69.5 & 58.2 & 86.8 & 66.2 & 100.0\% \\ 
    \midrule
    ToMe                        &   & \cmark & 64 &0.83 (10.9\%) &- & 43.7 & 38.9 & 48.6 & 57.1 & 50.0 & 45.3 & 52.5 & -    & 70.9\% \\
    PDrop                &   & \cmark & 64 &0.83 (10.9\%) & 1309.2 & 58.8 & 50.5 & 47.5 & 69.2 & 69.0 & 50.6 & 55.9 & -    & 85.0\% \\
    HiRED                      &   & \cmark & 64 &0.83 (10.9\%) & -      & 60.2 & 51.3 & 54.6 & 69.7 & 68.2 & 44.2 & 73.6 & -    & 88.2\% \\
    HiPrune                    &   & \cmark & 64 &0.83 (10.9\%) & -      & 59.5 & 53.4 & 53.6 & 69.2 & 68.9 & 54.9 & 73.0 & -    & 90.9\% \\
    FlowCut                    &   & \cmark & 64 &0.83 (10.9\%) & -      & 60.8 & 55.4 & 55.6 & 72.8 & 69.1 & 55.6 & 80.2 & -    & 94.2\% \\
    HoloV                      &   & \cmark & 64 &0.83 (10.9\%) & -      & 63.3 & 55.1 & 55.3 & 72.8 & 69.5 & 55.4 & 80.3 & -    & 94.6\% \\
    VISA                       &   & \cmark & 64 &0.83 (10.9\%) & 1420.6 & 62.1 & 57.3 & 56.2 & 74.1 & 67.9 & 55.6 & 77.6 & -    & 94.6\% \\
    D$^2$Pruner                &   & \cmark & 64 &0.83 (10.9\%) & -      & 61.9 & 55.6 & 57.9 & 74.6 & 70.0 & 56.1 & 82.4 & -    & 96.0\% \\
    VScan                      &   & \cmark & 64 &0.83 (10.9\%) & -      & 62.1 & 55.7 & 58.3 & 75.4 & 69.1 & 55.6 & 85.0 & -    & 96.4\% \\

    FiCoCo-L                   &   & \cmark & 58 &0.75 (9.9\%) & -      & 61.5 & 53.3 & 53.2 & 69.7 & 69.5 & 55.7 & 82.1 & -    & 93.1\% \\
    FastV                      &   & \cmark & 32 & 0.42 (5.4\%) & 884.6  & 37.8 & 33.2 & 41.5 & 43.4 & 42.6 & 42.5 & 32.5 & -    & 58.6\% \\
    SparseVLM                  &   & \cmark & 32 & 0.42 (5.4\%) & 1046.7 & 51.4 & 40.6 & 48.3 & 58.6 & 57.3 & 46.1 & 67.9 & -    & 76.4\% \\
    VisPruner                  &   & \cmark & 32 & 0.42 (5.4\%) & 1271.0 & 58.4 & 52.7 & 52.2 & 67.7 & 69.2 & 53.9 & 72.7 & 54.3 & 88.2\% \\
    DivPrune                   &   & \cmark & 32 & 0.42 (5.4\%) & 1284.9 & 57.6 & 49.1 & 54.9 & 71.2 & 68.6 & 52.9 & 81.5 & 58.7 & 90.0\% \\
    DOP\textsubscript{V}       & \cmark & \cmark & 32 & 0.42 (5.4\%) & 1306.5 & 59.4 & 53.7 & 54.8 & 71.0 & 69.1 & 54.5 & 79.6 & 56.7 & 91.2\% \\
    CDPruner                   &   & \cmark & 32 & 0.42 (5.4\%) & 1373.0 & 59.6 & 49.6 & 57.0 & 73.6 & 69.5 & 53.2 & 87.9 & -    & 93.2\% \\
    DOP\textsubscript{CD}      & \cmark & \cmark & 32 & 0.42 (5.4\%) & 1397.5 & 60.1 & 52.2 & 58.1 & 74.7 & 69.3 & 54.2 & 87.9 & 82.2 & 94.7\% \\

    \rowcolor{red!8}
    PDrop                       &   & \cmark & 270 & 3.54 (46.3\%)  & 1490.1 & 63.9 & 56.7 & 61.7 & 78.7 & 70.1 & 57.7 & 86.9 & 65.8 & 99.4\% \\
    \rowcolor{red!8}
    Dynamic-LLaVA               &   & \cmark & 115 & 1.50 (19.6\%) & 1479.8 & 65.4 & -    & 61.4 & 78.0 & 69.1 & 57.0 & 85.0 & 64.6 & 98.8\% \\
    \rowcolor{red!8}
    YOPO                        & \cmark &   & 70  & 0.92 (12.0\%) & 1423.5 & 64.6 & 57.0 & 60.7 & 77.4 & 68.0 & 55.2 & 86.6 & 64.6 & 97.7\% \\
    \rowcolor{red!8}
    TwigVLM                     &   & \cmark & 64  & 0.83 (10.9\%) & 1404.0 & 60.4 & 53.8 & 61.2 & 75.6 & 70.0 & 55.8 & 82.7 & 56.9 & 94.7\% \\
    \rowcolor{red!8}
    VisionZip                   &   & \cmark & 64  & 0.83 (10.9\%) & -      & 61.5 & -    & 57.0 & 74.2 & 68.8 & 56.0 & 80.9 & -    & 95.0\% \\
    \rowcolor{red!8}
    DART                        &   & \cmark & 64  & 0.83 (10.9\%) & -      & 64.7 & 56.7 & 57.1 & 74.6 & 71.1 & 54.7 & 79.3 & -    & 96.1\% \\
    \rowcolor{red!8}
    LLaVA-PruMerge              &   & \cmark & 32  & 0.42 (5.4\%)  & 1350.3 & 60.9 & 50.0 & 57.2 & 72.0 & 68.5 & 56.0 & 76.3 & 50.7 & 90.4\% \\
    \rowcolor{red!8}
    TRIM                        &   & \cmark & 29  & 0.38 (4.9\%)  & 1415.4 & 63.3 & 46.6 & 58.4 & 71.5 & 67.9 & 49.1 & 84.8 & 61.8 & 92.3\% \\
    \rowcolor{red!8}
    TokenPacker        &   & \cmark & 16  & 0.21 (2.7\%)  & 1378.8 & 62.7 & -    & 58.9 & 74.4 & 68.1 & 52.5 & 83.7 & -    & 94.7\% \\
    \rowcolor{red!8}
    Delta-LLaVA         &   & \cmark & 16  & 0.21 (2.7\%)  & 1375.9 & 62.9 & -    & 59.5 & 73.1 & 69.7 & 53.6 & 84.7 & -    & 95.4\% \\

    \rowcolor{lightblue}
    ViCA (ours)           & \cmark &   & 24  & 0.31 (4.1\%)   & 1464.5 & 64.0 & 57.7 & 60.4 & 76.6 & 68.5 & 55.5 & 86.7 & 63.2 & 97.8\% \\
    \rowcolor{lightblue}
    ViCA+PDrop${}^\dagger$ (ours) & \cmark & \cmark & 12  & 0.16 (2.0\%)   & 1449.7 & 63.3 & 56.9 & 59.0 & 75.8 & 69.3 & 54.4 & 85.2 & 60.8 & 96.3\% \\
    \bottomrule
    \end{tabular}
    }
\caption{ \small
Performance comparison of different pruning approaches on LLaVA-1.5-7B across nine benchmarks. We report per-benchmark accuracy and the average performance relative to the original model. Methods are grouped by strategy: unshaded for training-free, red for training-based, and blue for ours.  ViCA is retrained under the proposed minimal efficient architecture with the standard two-stage pipeline; ViCA+PDrop$^\dagger$ further applies PyramidDrop at inference. Op/Tok denote operation-/token-level pruning. Vision-related computation and token counts are measured following the FLOPs formulation in Appendix~\ref{sec:Calculation-Equation}.
}
\label{tab:finally-result}
\vspace{-10pt}
\end{table*}

\subsection{Setup}
\paragraph{Models.}
Within the LLaVA-1.5 architecture~\citep{liu2024improved}, we evaluate our approach using three different LLM backbones: MobileLLaMA-2.7B~\citep{wu2024mobilevlm}, Vicuna-7B-v1.5, and Vicuna-13B-v1.5~\citep{zheng2023judging}. In this paper, we denote the resulting models as LLaVA-1.5-3B, LLaVA-1.5-7B, and LLaVA-1.5-13B, respectively. Details are provided in Appendix~\ref{app:details_llm}. 

\paragraph{Benchmarks.}
We conduct experiments on nine standard image-based benchmarks: MME\textsuperscript{P}~\citep{fu2023mme}, MMB, MMB\textsuperscript{CN}~\citep{liu2024mmbench}, GQA~\citep{hudson2019gqa}, VQA\textsuperscript{v2}~\citep{goyal2017VQAv2}, SQA\textsuperscript{I}~\citep{lu2022ScienceQA}, VQA\textsuperscript{T}~\citep{singh2019TextVQA}, POPE~\citep{li-etal-2023-evaluating}, and SEED\textsuperscript{I}~\citep{li2024seed}. Details are in Appendix~\ref{app:benchmark}.

\paragraph{Implementation Details.}
We adopt LLaVA-1.5 as our experimental platform because it is a widely used open-source MLLM and a representative testbed for multimodal token compression. Its fully public training data and pipeline, together with a well-established two-stage recipe, enable reproducible and apples-to-apples comparisons. In the training-based setting, we follow the official data and two-stage protocol~\citep{liu2024improved}, train from scratch with reduced visual computation, and keep the iterations identical to the original model. All experiments are conducted on A100 GPUs.

\paragraph{Efficiency Evaluation.}
For theoretical efficiency, we report vision-side TFLOPs and FLOPs-equivalent visual token counts (Appendix~\ref{para:equiv-token-count}). For practical efficiency, we benchmark inference on 5,000 TextVQA samples using an NVIDIA A6000 GPU. Since both the baseline and our method achieve near-saturated CUDA utilization, the observed speedups mainly reflect computation reduction rather than hardware underutilization. We report CUDA utilization and prefill latency, which dominates inference cost.

\vspace{-2pt}
\subsection{Main Results}

\paragraph{Task Performance.}
As shown in Table~\ref{tab:finally-result}, ViCA retains 97.8\% relative accuracy with only 4.1\% vision-side computation; with inference-time token compression via PyramidDrop, ViCA+PDrop further reduces computation to 2.0\% while retaining 96.3\% relative accuracy, outperforming existing methods at comparable or larger budgets. Similar trends hold for LLaVA-1.5-3B and 13B (Appendix Table~\ref{tab:llava15_3b_13b_merged}), where ViCA and ViCA+PDrop consistently maintain strong relative accuracy under extremely small vision-compute budgets. Beyond standard benchmarks, ViCA also maintains favorable performance on vision-heavy and spatial grounding tasks, suggesting that the reduced visual computation still preserves fine-grained visual understanding (Appendix Tables~\ref{tab:vision_heavy}--\ref{tab:grounding}). Overall, ViCA achieves a favorable accuracy--efficiency trade-off across model scales and task types.

\paragraph{Complementarity with Token Pruning.}
Our approach removes redundant visual-token update paths architecturally, making ViCA orthogonal to token-dropping methods. As a result, ViCA can be seamlessly combined with PyramidDrop at inference time without modifying the training procedure. This combination further reduces vision-related computation to 1.7\%--2.0\% across LLaVA-1.5-3B/7B/13B, while incurring only a modest 1.8\%--3.7\% accuracy drop relative to the original full models. These results confirm that architectural-level visual compute removal and token-level pruning are complementary. Implementation details of PyramidDrop are provided in Appendix~\ref{sec:pdrop}.

\begin{table}[t]
\centering
\scriptsize
\setlength{\tabcolsep}{2.2pt}
\renewcommand{\arraystretch}{0.95}
\resizebox{\columnwidth}{!}{
\begin{tabular}{cc|cccc|c}
\toprule
\textbf{Base Model} & \textbf{Method} &
\textbf{\makecell[c]{Vision\\Tokens}} &
\textbf{\makecell[c]{Vision\\TFLOPs}} &
\textbf{\makecell[c]{Total\\TFLOPs}} &
\textbf{\makecell[c]{Vision\\KV-Cache}} &
\textbf{\makecell[c]{Rel.\\Avg.}} \\
\midrule

\multirow{4.5}{*}{\makecell[c]{LLaVA-1.5-3B}}
& Original      & 576 & 3.04 & 3.33 & 100.0\% & 100.0\% \\
& PDrop         & 270 & 1.40 & 1.68 & 46.9\%  & 97.0\% \\
\cmidrule(lr){2-7}
& ViCA          & 27  & 0.14\(\downarrow\)95.5\%  & 0.42 & 28.1\% & 98.7\% \\
& ViCA+PDrop${}^\dagger$    & 12  & 0.06\(\downarrow\)98.0\% & 0.34 & 12.5\% & 98.2\% \\
\midrule
\multirow{4.5}{*}{LLaVA-1.5-7B}
& Original      & 576 & 7.65 & 8.38 & 100.0\% & 100.0\% \\ 
& PDrop         & 270 & 3.54 & 4.26 & 46.9\%  & 99.4\% \\  
\cmidrule(lr){2-7}
& ViCA          & 24  & 0.31\(\downarrow\)95.9\% & 1.02 & 25.0\% & 97.8\% \\
& ViCA+PDrop${}^\dagger$    & 12  & 0.16\(\downarrow\)98.0\% & 0.87 & 12.5\% & 96.3\% \\
\midrule
\multirow{4.5}{*}{LLaVA-1.5-13B}
& Original      & 576 & 14.91 & 16.34 & 100.0\% & 100.0\% \\
& PDrop${}^\dagger$         & 270 & 6.92 & 8.33 & 46.9\%  & 99.8\% \\
\cmidrule(lr){2-7}
& ViCA          & 19  & 0.49\(\downarrow\)96.7\% & 1.88 & 20.0\% & 97.0\% \\
& ViCA+PDrop${}^\dagger$    & 10  & 0.26\(\downarrow\)98.3\% & 1.66 & 10.6\% & 96.3\% \\
\bottomrule
\end{tabular}
}
\caption{\small Theoretical efficiency comparison across three LLM backbones within the LLaVA-1.5 framework.}
\label{tab:theoretical}
\vspace{-10pt}
\end{table}

\paragraph{Theoretical Efficiency.}
As shown in Table~\ref{tab:theoretical}, under the ViCA architecture, the vision-related computation of LLaVA-1.5-3B/7B/13B is reduced to 4.5\%/4.1\%/3.3\% of the original models. Under the FLOPs-based analysis, this reduction is equivalent to compressing the number of visual tokens from 576 to 27/24/19, respectively. Correspondingly, the visual KV-cache sizes are reduced to 28.1\%/25.0\%/20.0\% of the original. When combined with PyramidDrop at inference time, the vision-related computation is further reduced to 2.0\%/2.0\%/1.7\%, corresponding to 12/12/10 equivalent visual tokens, with the visual KV-cache sizes reduced to 12.5\%/12.5\%/10.6\% of the original.

\begin{table}[t]
\centering
\resizebox{\columnwidth}{!}{
\begin{tabular}{lccccc}
\toprule
\textbf{Base Model} & 
\textbf{Method} & 
\textbf{\makecell[c]{CUDA\\Util}} &
\textbf{\makecell[c]{Latency. \\(SpdUp)}} &
\textbf{\makecell[c]{Latency Vis.\\(Rel.)}} \\
\midrule
\multirow{3}{*}{LLaVA-1.5-3B}
& Original                      & 78\% & 57.9 (1.0$\times$) & 35.6 (100.0\%) \\
& w/o Vis. Tok.                 & 70\% & 22.3 (2.6$\times$) & 0.0 (0.0\%) \\
& Ours                          & 65\% & 24.2 (2.4$\times$) & 1.9 (5.3\%) \\
\midrule
\multirow{3}{*}{LLaVA-1.5-7B}
& Original                      & 97\% & 124.5 (1.0$\times$) & 94.2 (100.0\%) \\
& w/o Vis. Tok.                 & 95\% & 30.3 (4.1$\times$)  & 0.0 (0.0\%) \\
& Ours                          & 94\% & 36.0 (3.5$\times$)  & 5.7 (6.1\%) \\
\midrule
\multirow{3}{*}{LLaVA-1.5-13B}
& Original                      & 97\% & 210.6(1.0$\times$) & 159.3 (100.0\%) \\
& w/o Vis. Tok.                 & 90\% & 51.3(4.1$\times$)  & 0.0 (0.0\%) \\
& Ours                          & 98\% & 59.3(3.6$\times$)  & 8.0 (5.0\%) \\
\bottomrule
\end{tabular}
}
\caption{\small 
Practical efficiency under FlashAttention across three LLM backbones in LLaVA-1.5. We report efficiency for the baseline, text-only setting, and our method, averaged over 5,000 TextVQA samples on an NVIDIA A6000 GPU.
}
\label{tab:practical}
\vspace{-5pt}
\end{table}

\begin{figure}[t]
\centering
\includegraphics[width=\linewidth, trim=0 9 0 0, clip]{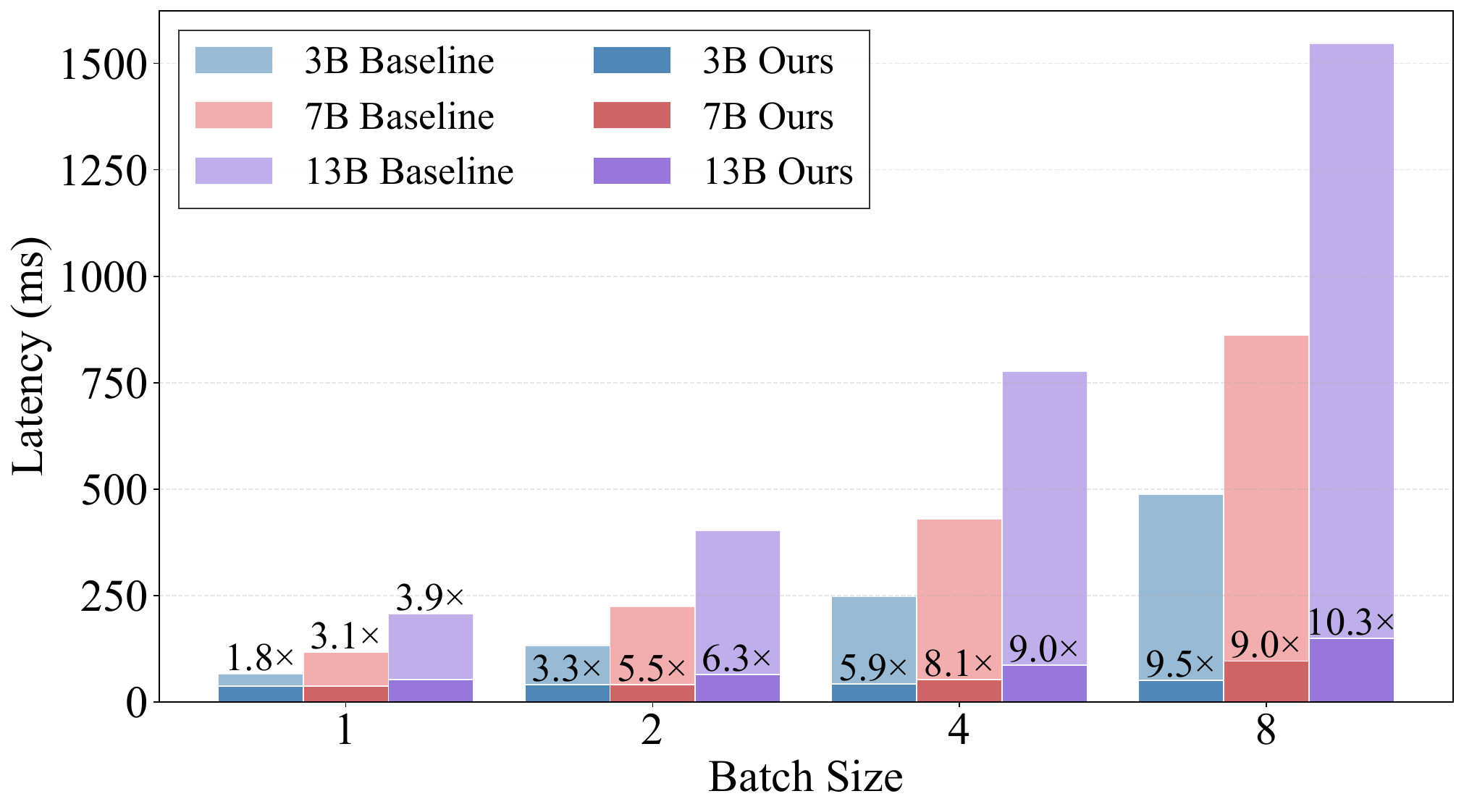}
\caption{\small 
Prefill latency and speedup across batch sizes on an A6000 GPU. Results are averaged over 100 samples, with speedups annotated for LLaVA-1.5-3B, 7B, and 13B.
}
\label{fig:latency_batch_size}
\vspace{-5pt}
\end{figure}

\paragraph{Practical Efficiency.}
In multimodal inference, the prefill stage dominates computation and end-to-end latency because it processes all visual tokens and incurs most vision-related overhead. We therefore focus our analysis on prefill performance. As shown in Table~\ref{tab:practical}, ViCA achieves prefill latency close to that of the text-only model, yielding speedups of 2.4$\times$ / 3.5$\times$ / 3.6$\times$ over the original LLaVA-1.5 3B / 7B / 13B models. Figure~\ref{fig:latency_batch_size} further shows that these gains persist across batch sizes, with stable speedup trends and larger absolute latency reductions for larger models. The smaller speedup on the 3B model mainly results from reduced CUDA utilization. As visual computation is aggressively removed, fewer streaming multiprocessors (SMs) are scheduled, leaving some previously active SMs underutilized. In contrast, the 7B and 13B models operate near saturation and translate computation reduction more effectively into latency gains. Finally, isolating the latency introduced by visual computation by subtracting the text-only prefill latency shows that ViCA accounts for only 5.0\%--6.1\% of the original visual latency. This closely matches the theoretical visual computation ratios of 3.3\%--4.5\% in Table~\ref{tab:practical}, validating our FLOPs-based analysis. We measure the prefill forward-pass latency across batch sizes to simulate multi-batch inference. As shown in Figure~\ref{fig:latency_batch_size}, at batch size 8, ViCA achieves approximately 10$\times$ speedup over the original LLaVA-1.5 models.

\begin{table*}[t!]
\centering
\resizebox{\textwidth}{!}{
\begin{tabular}{c|cc|ccccccccc|c}
\toprule
\multirow{2}{*}{\textbf{\#}} &
\multicolumn{2}{c|}{\textbf{Frozen Update}} &
\multirow{2}{*}{\textbf{MME\textsuperscript{P}}} &
\multirow{2}{*}{\textbf{MMB}} &
\multirow{2}{*}{\textbf{MMB\textsuperscript{CN}}} &
\multirow{2}{*}{\textbf{GQA}} &
\multirow{2}{*}{\textbf{VQA\textsuperscript{v2}}} &
\multirow{2}{*}{\textbf{SQA\textsuperscript{I}}} &
\multirow{2}{*}{\textbf{VQA\textsuperscript{T}}} &
\multirow{2}{*}{\textbf{POPE}} &
\multirow{2}{*}{\textbf{SEED\textsuperscript{I}}} &
\multirow{2}{*}{\textbf{Rel. Avg.}} \\
  & \textbf{Vis. Attn} & \textbf{Vis. FFN} & &  &  &  & &  &  & & & \\
\midrule
1 & \xmark  & \xmark  & 1506.5 & 64.7 & 58.1 & 61.9 & 78.5 & 69.5 & 58.2 & 86.8 & 66.2 & 100.0\% \\
2 &\cmark   & \xmark & 1501.1 & 66.1 & 59.5 & 62.1 & 78.0 & 69.3 & 57.1 & 87.1 & 65.8 & 100.2\% \\ 
3 &\xmark & \cmark   & 1472.4 & 65.6 & 56.9 & 62.0 & 77.5 & 70.3 & 56.9 & 87.3 & 64.6 & 99.2\%  \\
4 & \cmark & \cmark & 1471.1 & 65.2 & 56.8 & 61.8 & 77.2 & 69.8 & 56.6 & 86.8 & 64.0 & 98.8\%  \\
\bottomrule
\end{tabular}
}
\caption{\small 
Performance comparison on LLaVA-1.5-7B under different visual token update freezing strategies Results across nine benchmarks are reported for three freezing variants, evaluated after standard two-stage retraining.
}
\label{tab:7b_ablation}
\vspace{-5pt}
\end{table*}

\begin{table*}[t]
\centering
\resizebox{0.98\textwidth}{!}{
\begin{tabular}{cc|ccccccccc|cc}
\toprule
\textbf{Base Model} & \textbf{Method} &
\textbf{MME\textsuperscript{P}} &
\textbf{MMB} &
\textbf{MMB\textsuperscript{CN}} &
\textbf{GQA} &
\textbf{VQA\textsuperscript{v2}} &
\textbf{SQA\textsuperscript{I}} &
\textbf{VQA\textsuperscript{T}} &
\textbf{POPE} &
\textbf{SEED\textsuperscript{I}} &
\textbf{Rel. Avg.} &
\textbf{Vis. FLOPs} \\
\midrule
\multirow{3}{*}{LLaVA-1.5-3B}
& Original  & 1258.2 & 57.0 & 28.6 & 59.4 & 75.3 & 60.0 & 48.6 & 86.4 & 60.1 & 100.0\% & 100.0\% \\
& FreezeVis  & 1212.0 & 54.8 & 34.9 & 57.5 & 72.8 & 61.2 & 46.8 & 86.5 & 57.8 & 100.3\% & 16.0\% \\
& ViCA & 1188.9 &55.4 &33.2 &56.6 &72.1 & 61.2 &45.4 &86.2 &56.7 & 98.7\% & 4.5\% \\
\midrule
\multirow{3}{*}{LLaVA-1.5-7B}
& Original  & 1506.5 & 64.7 & 58.1 & 61.9 & 78.5 & 69.5 & 58.2 & 86.8 & 66.2 & 100.0\% & 100.0\% \\
& FreezeVis  & 1471.1 & 65.2 & 56.8 & 61.8 & 77.2 & 69.8 & 56.6 & 86.8 & 64.0 & 98.8\% & 16.2\% \\
& ViCA & 1464.5 & 64.0 & 57.7 & 60.4 & 76.6 & 68.5 & 55.5 & 86.7 & 63.2 & 97.8\% & 4.1\% \\
\midrule
\multirow{3}{*}{LLaVA-1.5-13B}
& Original  & 1529.9 & 68.5 & 63.5 & 63.3 & 80.0 & 72.8 & 61.2 & 87.0 & 68.2 & 100.0\% & 100.0\% \\
& FreezeVis  & 1474.5 & 67.3 & 60.7 & 62.4 & 78.2 & 71.0 & 58.2 & 86.9 & 66.2 & 97.4\% & 16.3\% \\
& ViCA & 1453.6 &67.1 &62.3 &61.8 &77.9 &72.2 &56.7 &86.5 &65.2 & 97.0\% & 3.3\% \\
\bottomrule
\end{tabular}
}
\caption{\small 
Performance comparison across LLaVA-1.5-3B, 7B, and 13B under different architectural variants.
Performance of architectural variants across LLaVA-1.5-3B/7B/13B. We compare dense self-attention (Original), dense cross-attention with frozen visual-token updates (FreezeVis), and our minimal efficient architecture with selected-layer sparse cross-attention (ViCA). All variants use the standard two-stage training protocol and are evaluated on nine benchmarks. FLOPs follow Appendix~\ref{sec:Calculation-Equation}.
}
\label{tab:all_ablation}
\vspace{-10pt}
\end{table*}

\paragraph{Parallel Decoupling.}
Under ViCA, vision-side computation is confined to two components: linear projections that map visual tokens to key--value representations and the matrix multiplications in text--vision cross-attention. For LLaVA-1.5-7B, the computation ratio between visual key--value projection and cross-attention is approximately 204:1 (Appendix~\ref{sec:Calculation-Equation}). By freezing visual tokens, ViCA enables parallel decoupling, allowing raw visual tokens to be directly fed into the key--value projection module and precomputed in parallel with the Transformer backbone. As a result, the remaining $\sim$4\% vision-side computation can be parallelized, effectively reducing visual latency to near zero.

\subsection{Ablation Studies}
\label{sec:ablation_studies}
\paragraph{Freezing Visual Token Updates.}
We first analyze LLaVA-1.5-7B by independently freezing the attention and FFN update paths of visual tokens. As shown in Table~\ref{tab:7b_ablation}, freezing attention updates improves relative average accuracy by 0.2\%, whereas freezing FFN updates results in a 0.8\% drop, indicating that attention-based visual-token updates are less critical than FFN-based updates. We then freeze both update paths across LLaVA-1.5-3B, 7B, and 13B models. As summarized in Table~\ref{tab:all_ablation}, the resulting models retain 100.3\%, 98.8\%, and 97.4\% of their  baseline accuracies, while reducing visual computation to approximately 16\% of the original, demonstrating that disabling visual-token updates largely preserves performance while substantially reducing visual computation.

\paragraph{Key Cross-attention Layer Selection.}
As shown in Table~\ref{tab:all_ablation}, disabling visual-token updates while retaining dense cross-attention (FreezeVis) preserves nearly full accuracy, and further restricting cross-attention to a small set of diagnostically identified key layers (ViCA) incurs only a 1--3\% accuracy drop while reducing vision-related FLOPs from approximately 16\% to 4\%. These results show that visual tokens need not be continuously updated across all Transformer layers and that multimodal reasoning relies on only a small subset of critical text--vision cross-attention layers, supporting the proposed minimal and efficient architecture.

%% file: Sec/7_conclusion.tex
\section{Conclusion}
\label{sec:conclusion}
\vspace{-2pt}
In this work, we conduct a systematic architectural diagnosis of self-attention LVLMs and find that continuously updating visual tokens via self-attention and FFN layers is largely redundant, while effective cross-modal interaction is concentrated in a small subset of critical layers. Based on this insight, we propose ViCA, which removes redundant visual computation paths and yields a more regular, hardware-friendly inference pipeline, achieving approximately $3.5\times$ prefill speedup  in single-batch inference and $>10\times$ speedup in multi-batch inference. The resulting minimal efficient architecture is also naturally compatible with existing token dropping methods, enabling further efficiency gains when combined. Looking forward, we will explore adaptive architectures for improved cross-modal fusion and aim to unify diverse pruning paradigms into a general framework for efficient and scalable multimodal architectures.

%% file: Sec/8_limitation.tex
\section*{Limitations}
\label{sec:limitation}

This work leaves two important directions for future exploration. First, our evaluation is conducted within the LLaVA-1.5 framework, whose training data and two-stage training pipeline are fully open and widely adopted, enabling controlled comparisons for training-based efficiency methods. Since ViCA is a training-based architectural compression method, extending it to more recent MLLMs such as LLaVA-OneVision, Qwen-VL, or InternVL is less straightforward: their complete training data, data mixtures, or training recipes are not always publicly available, and reproducing them often requires substantially larger-scale training. These factors make fully controlled evaluation challenging, so we leave large-scale validation on these models to future work. Second, this paper focuses on the single-image setting. Our results show that visual-token write operations are largely redundant in the evaluated single-image MLLMs. However, multi-image and video scenarios may require richer vision-to-vision interactions to capture cross-image correspondence, temporal continuity, and fine-grained visual grounding. Extending ViCA to these settings, potentially with adaptive visual interaction modules while preserving efficient text--vision interaction, remains an important direction for future work.

%% file: Sec/x_supp.tex

\section{The Use of Large Language Models}

We employed large language models (LLMs) solely as general-purpose writing assistants for language refinement, including improving clarity, grammar, and style.
Importantly, no LLM was involved in research ideation, methodological design, analysis, or result interpretation; the role of the LLM was limited to linguistic polishing.
All substantive contributions originated from the authors.
This ensured that the scientific content remained entirely authored by the researchers, while benefiting from improved academic writing quality.

\section{Additional Results}
\label{app:add_exp}

\subsection{Cross-Scale Generalization}
\label{app:cross_scale}
To assess cross-scale generalization, we report results on LLaVA-1.5-3B and LLaVA-1.5-13B in Table~\ref{tab:llava15_3b_13b_merged}, comparing our method with representative training-free and training-based pruning approaches. On LLaVA-1.5-3B, our method preserves 98.7\% of baseline accuracy while reducing vision-side computation to 4.5\%, which can be further lowered to 2.0\% when combined with PyramidDrop at inference time. On LLaVA-1.5-13B, it achieves 97.0\% relative accuracy using only 3.3\% of the original visual computation, further reduced to 1.7\% with PyramidDrop. These results demonstrate that the proposed ViCA generalizes well across model scales, consistently achieving a strong accuracy-efficiency trade-off.

\begin{table*}[htbp]
\centering
\resizebox{0.97\textwidth}{!}{
\begin{tabular}{c|cc|cc|ccccccccc|c}
\toprule
\multirow{2}{*}{\textbf{Method}} &
\multicolumn{2}{c|}{\textbf{Sparsity}} &
\multicolumn{2}{c|}{\textbf{Vision-side}} &
\multirow{2}{*}{\textbf{MME\textsuperscript{P}}} &
\multirow{2}{*}{\textbf{MMB}} &
\multirow{2}{*}{\textbf{MMB\textsuperscript{CN}}} &
\multirow{2}{*}{\textbf{GQA}} &
\multirow{2}{*}{\textbf{VQA\textsuperscript{v2}}} &
\multirow{2}{*}{\textbf{SQA\textsuperscript{I}}} &
\multirow{2}{*}{\textbf{VQA\textsuperscript{Text}}} &
\multirow{2}{*}{\textbf{POPE}} &
\multirow{2}{*}{\textbf{SEED\textsuperscript{I}}} &
\multirow{2}{*}{\textbf{Rel. Avg.}} \\
& \textbf{Op.} & \textbf{Tok.} & \textbf{Token} & \textbf{TFLOPs (Rel.)} &  \\
\midrule
\multicolumn{15}{l}{\textbf{LLaVA-1.5-3B}} \\
\midrule
Original & - & - & 576 & 3.04 (100.0\%) & 1258.2 & 57.0 & 28.6 & 59.4 & 75.3 & 60.0 & 48.6 & 86.4 & 60.1 & 100.0\% \\
\midrule
\rowcolor{red!8}
PDrop &   & \cmark & 270 & 1.40 (46.0\%) & 1231.1 & 54.3 & 29.0 & 57.0 & 72.9 & 60.3 & 48.7 & 87.1 & 61.4 & 97.0\% \\
\rowcolor{red!8}
YOPO & \cmark &   & 70 & 0.37 (12.0\%) & 1106.5 & 48.6 & 29.0 & 50.4 & 64.3 & 56.5 & 39.6 & 83.7 & 51.4 & 89.2\% \\
\midrule
\rowcolor{lightblue}
ViCA & \cmark &   & 27 & 0.14 (4.5\%) & 1188.9 & 55.4 & 33.2 & 56.6 & 72.1 & 61.2 & 45.4 & 86.2 & 56.7 & 98.7\% \\
\rowcolor{lightblue}
ViCA+PDrop${}^\dagger$ & \cmark & \cmark & 12 & 0.06 (2.0\%) & 1187.5 & 56.3 & 34.5 & 54.8 & 71.1 & 60.8 & 44.8 & 84.3 & 55.2 & 98.2\% \\
\midrule
\multicolumn{15}{l}{\textbf{LLaVA-1.5-13B}} \\
\midrule
Original & - & - & 576 & 14.91 (100.0\%) & 1529.9 & 68.5 & 63.5 & 63.3 & 80.0 & 72.8 & 61.2 & 87.0 & 68.2 & 100.0\% \\
\midrule
FastV &   & \cmark & 64 & 1.63 (10.9\%) & 1246.4 & 59.2 & 55.1 & 51.9 & 65.3 & 73.1 & 53.4 & 56.9 & 56.1 & 83.7\% \\
ToMe &   & \cmark & 64 & 1.63 (10.9\%) & 1287.0 & 59.5 & 51.4 & 55.3 & 67.1 & 70.7 & 50.1 & 71.5 & -    & 85.5\% \\
PDrop &   & \cmark & 64 & 1.63 (10.9\%) & 1247.0 & 63.1 & 56.6 & 54.1 & 70.8 & 73.1 & 55.3 & 66.1 & 58.4 & 87.7\% \\
HiPrune &   & \cmark & 64 & 1.63 (10.9\%) & -      & 64.8 & 59.2 & 54.2 & 70.3 & 74.6 & 56.7 & 72.4 & -    & 91.4\% \\
SparseVLM &   & \cmark & 64 & 1.63 (10.9\%) & 1374.3 & 65.2 & 60.3 & 55.9 & 73.2 & 73.0 & 57.1 & 77.9 & 60.3 & 92.4\% \\
VISA &   & \cmark & 64 & 1.63 (10.9\%) & 1468.3 & 66.0 & 60.3 & 57.4 & 75.5 & 73.9 & 59.2 & 79.5 & -    & 95.3\% \\
VisionZip &   & \cmark & 32 & 0.81 (5.5\%) & 1257.7 & 61.2 & 55.8 & 52.7 & 68.4 & 72.9 & 55.2 & 66.8 & 56.2 & 86.4\% \\
DART &   & \cmark & 32 & 0.81 (5.5\%) & 1282.8 & 61.9 & 56.2 & 53.9 & 68.1 & 73.2 & 55.1 & 66.9 & 57.6 & 87.2\% \\
VisPruner &   & \cmark & 32 & 0.81 (5.5\%) & 1314.2 & 61.3 & 56.1 & 52.6 & 69.0 & 71.7 & 56.0 & 71.9 & 56.5 & 87.6\% \\
DivPrune &   & \cmark & 32 & 0.81 (5.5\%) & 1405.2 & 61.7 & 57.2 & 56.2 & 72.0 & 70.9 & 54.6 & 79.3 & 60.1 & 90.7\% \\
DOP\textsubscript{V} & \cmark & \cmark & 32 & 0.81 (5.5\%) & 1365.6 & 64.3 & 59.5 & 56.0 & 73.2 & 73.6 & 57.2 & 78.8 & 59.7 & 92.2\% \\
CDPruner &   & \cmark & 32 & 0.81 (5.5\%) & 1421.0 & 63.7 & 56.6 & 58.5 & 75.2 & 71.9 & 55.3 & 87.6 & 62.5 & 93.7\% \\
DOP\textsubscript{CD} & \cmark & \cmark & 32 & 0.81 (5.5\%) & 1468.9 & 65.1 & 58.7 & 59.0 & 76.2 & 68.9 & 56.7 & 87.5 & 63.8 & 94.8\% \\
\rowcolor{red!8}
PDrop &   & \cmark & 270 & 6.92 (46.4\%) & 1555.2 & 68.8 & 63.6 & 63.1 & 79.6 & 70.9 & 60.8 & 87.6 & 68.1 & 99.8\% \\
\rowcolor{red!8}
Delta-LLaVA &   & \cmark & 144 & 3.68 (24.7\%) & 1527.5 & 67.4 & -    & 62.7 & 76.9 & 71.6 & 59.2 & 87.3 & -    & 98.4\% \\
\rowcolor{red!8}
Dynamic-LLaVA &   & \cmark & 115 & 2.93 (19.7\%) & 1554.1 & 68.3 & -    & 62.7 & 79.1 & 72.2 & 59.5 & 86.8 & 66.6 & 99.1\% \\
\rowcolor{red!8}
YOPO & \cmark &   & 70  & 1.78 (12.0\%) & 1430.0 & 66.2 & 60.6 & 61.2 & 78.1 & 70.7 & 56.7 & 86.4 & 65.4 & 96.1\% \\
\rowcolor{red!8}
VisionZip &   & \cmark & 64  & 1.63 (10.9\%) & -      & 65.5 & -    & 58.1 & 75.2 & 72.3 & 58.5 & 81.8 & 61.4 & 94.4\% \\
\rowcolor{red!8}
TwigVLM &   & \cmark & 64  & 1.63 (10.9\%) & 1503.1 & 66.2 & 61.1 & 62.5 & 77.2 & 73.6 & 59.5 & 84.1 & 63.2 & 97.1\% \\
\rowcolor{red!8}
LLaVA-PruMerge &   & \cmark & 32  & 0.81 (5.5\%) & 1428.6 & 62.3 & 54.5 & 53.3 & 72.8 & 71.0 & 58.4 & 78.5 & 54.4 & 89.8\% \\
\rowcolor{red!8}
TRIM &   & \cmark & 29  & 0.74 (4.9\%) & 1337.9 & 65.5 & 52.4 & 56.0 & 75.4 & 70.1 & 50.7 & 85.2 & 60.8 & 90.5\% \\
\rowcolor{lightblue}
ViCA & \cmark &   & 19 & 0.49 (3.3\%) & 1453.6 & 67.1 & 62.3 & 61.8 & 77.9 & 72.2 & 56.7 & 86.5 & 65.2 & 97.0\% \\
\rowcolor{lightblue}
ViCA+PDrop${}^\dagger$ & \cmark & \cmark & 10 & 0.26 (1.7\%) & 1442.1 & 65.7 & 62.7 & 60.7 & 77.4 & 72.1 & 56.8 & 85.9 & 64.7 & 96.3\% \\
\bottomrule
\end{tabular}
}
\caption{\small
Performance comparison of different pruning approaches on LLaVA-1.5-3B/13B across nine benchmarks.
}
\label{tab:llava15_3b_13b_merged}
\end{table*}

\subsection{Fine-Grained Tasks}
\label{app:fine_grained}
To further examine the impact of vision-side compression on fine-grained visual understanding, we evaluate ViCA on both vision-heavy benchmarks, including OCRBench~\citep{liu2024ocrbench}, BLINK~\citep{fu2024blink}, V\textsuperscript{*}~\citep{wu2024v}, AI2D~\citep{kembhavi2016diagram}, RealWorldQA~\citep{xai2024grok15v}, MMStar~\citep{chen2024we}, InfoVQA~\citep{mathew2022infographicvqa}, and Flickr30k-CIDEr~\citep{young2014image, vedantam2015cider}, and spatial grounding benchmarks, including RefCOCO, RefCOCO+~\citep{kazemzadeh2014referitgame,yu2016modeling}, and RefCOCOg~\citep{mao2016generation,yu2016modeling}. These tasks require accurate spatial localization, text recognition, and detailed visual perception, and therefore provide a direct test of whether aggressive visual token pruning harms fine-grained reasoning. More details of these benchmarks are provided in Appendix~\ref{app:benchmark}. 

\begin{table*}[t]
\centering
\resizebox{1.0\linewidth}{!}{
\begin{tabular}{c|cccccccc|cc}
\toprule
Method  & OCRBench & BLINK & V\textsuperscript{*} & AI2D & RealworldQA & MMStar & InfoVQA & Flickr30k-CIDEr & Rel. Avg & Vis. FLOPs\\
\midrule
Original & 31.4 & 38.7 & 50.3 & 55.1 & 55.8 & 33.7 & 20.4 & 74.9 & 100.0\% & 100.0\% \\
\midrule
PDrop    & 26.1 & 37.8 & 43.5 & 53.2 & 50.0 & 32.8 & 18.8 & 65.7 & 91.3\% & 11.1\%\\
ViCA     & \textbf{28.7} & \textbf{37.0} & \textbf{46.6} & \textbf{51.6} & \textbf{51.2} & \textbf{33.3} & \textbf{19.4} & \textbf{71.5} & \textbf{94.3\%} & \textbf{4.1\%} \\
\bottomrule
\end{tabular}
}
\caption{\small Performance comparison of different pruning approaches on LLaVA-1.5-7B across eight vision-heavy benchmarks.}
\label{tab:vision_heavy}
\end{table*}

\begin{table*}[t!]
\centering
\resizebox{1.0\linewidth}{!}{
\begin{tabular}{c|cc|cc|cc|cc|cc|cc}
\toprule
\multirow{2}{*}{\textbf{Method}} &
\multicolumn{2}{c|}{\textbf{RefCOCO\textsuperscript{testA}}} &
\multicolumn{2}{c|}{\textbf{RefCOCO\textsuperscript{testB}}} &
\multicolumn{2}{c|}{\textbf{RefCOCOg}} &
\multicolumn{2}{c|}{\textbf{RefCOCO+\textsuperscript{testA}}} &
\multicolumn{2}{c|}{\textbf{RefCOCO+\textsuperscript{testB}}} &
\multirow{2}{*}{\textbf{Rel. Avg}} &
\multirow{2}{*}{\textbf{Vis. FLOPs}} \\
& \textbf{Acc@0.5} & \textbf{Center Acc} & \textbf{Acc@0.5} & \textbf{Center Acc} & \textbf{Acc@0.5} & \textbf{Center Acc} & \textbf{Acc@0.5} & \textbf{Center Acc} & \textbf{Acc@0.5} & \textbf{Center Acc} & & \\
\midrule
Original    & 63.14 & 92.01 & 45.79 & 86.44 & 47.19 & 84.97 & 57.82 & 86.92 & 37.86 & 74.23 & 100.0\% & 100.0\% \\
\midrule
PDrop       & 28.32 & 71.06 & 21.04 & 70.93 & 18.15 & 59.63 & 22.30 & 56.93 & 14.79 & 50.19 & 57.0\% & 11.1\% \\
ViCA        & \textbf{40.91} & \textbf{81.54} & \textbf{30.66} & \textbf{77.94} & \textbf{28.00} & \textbf{73.21} & \textbf{34.65} & \textbf{73.56} & \textbf{24.18} & \textbf{62.24} & \textbf{74.8\%} & \textbf{4.1\%} \\

\bottomrule
\end{tabular}
}
\caption{\small Performance comparison of different pruning approaches on LLaVA-1.5-7B across three grounding benchmarks.}

\label{tab:grounding}
\end{table*}
As shown in Table~\ref{tab:vision_heavy}, on vision-heavy benchmarks, ViCA achieves higher relative performance than PDrop while using substantially less vision-side computation: 94.3\% vs. 91.3\% relative average performance, with 4.1\% vs. 11.1\% vision FLOPs. This suggests that removing redundant visual update paths better preserves vision-intensive reasoning than directly dropping visual tokens under aggressive compression.

Table~\ref{tab:grounding} further shows that ViCA maintains stronger spatial grounding performance than PDrop, improving the relative average from 57.0\% to 74.8\% while using only 4.1\% vision FLOPs. These results indicate that preserving visual information at critical cross-attention layers is important for object localization and fine-grained spatial reasoning. Although a gap remains compared with the original model, especially on grounding tasks, ViCA achieves a more favorable performance--efficiency trade-off across both vision-heavy and spatial grounding benchmarks.

\section{More Models Analysis}
\label{app:discussion}

\begin{figure*}[htbp]
\centering
\begin{subfigure}[t]{0.32\linewidth}
    \centering
    \includegraphics[width=\linewidth]{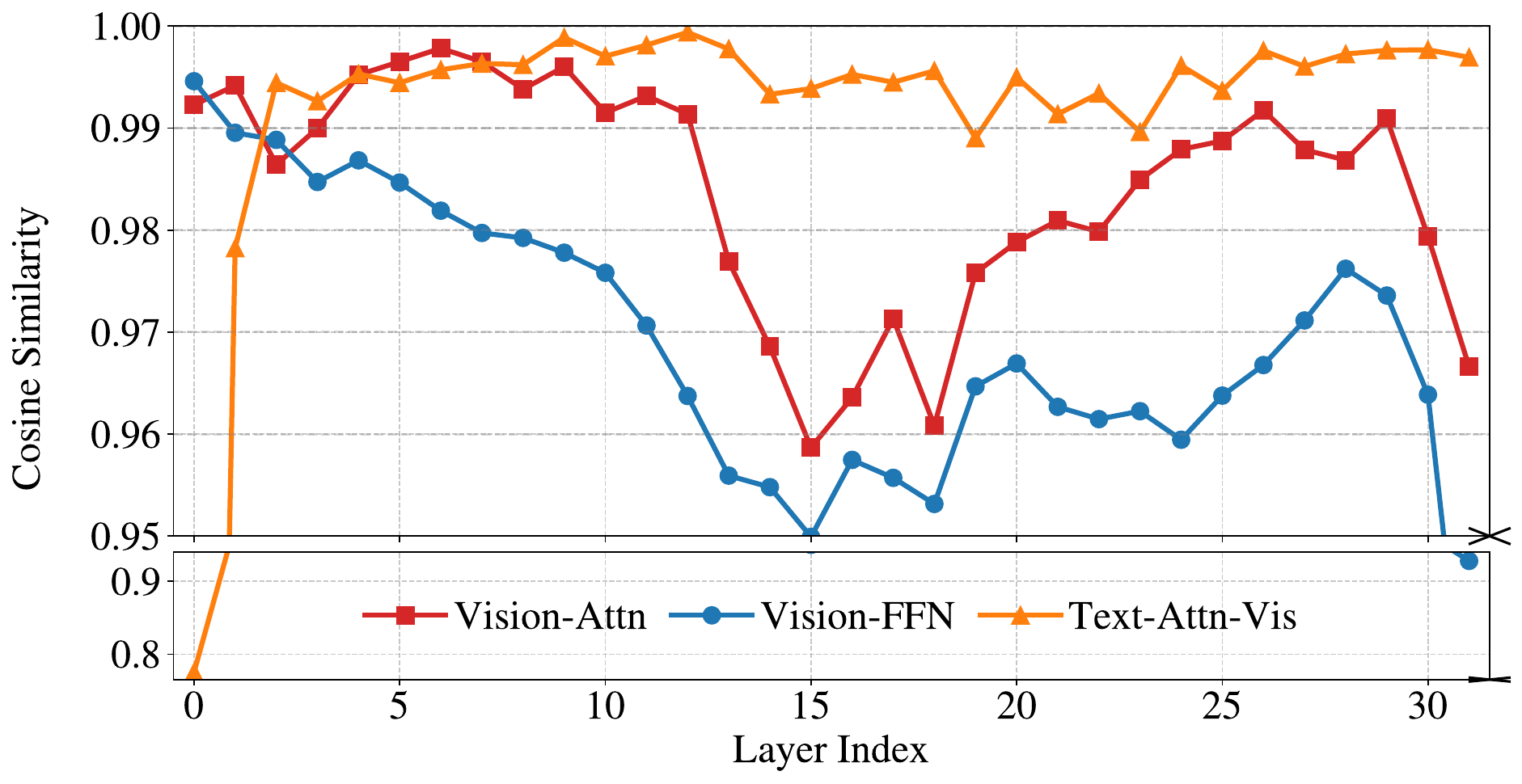}
    \caption{LLaVA-1.5-3B}
\end{subfigure}
\hfill
\begin{subfigure}[t]{0.32\linewidth}
    \centering
    \includegraphics[width=\linewidth]{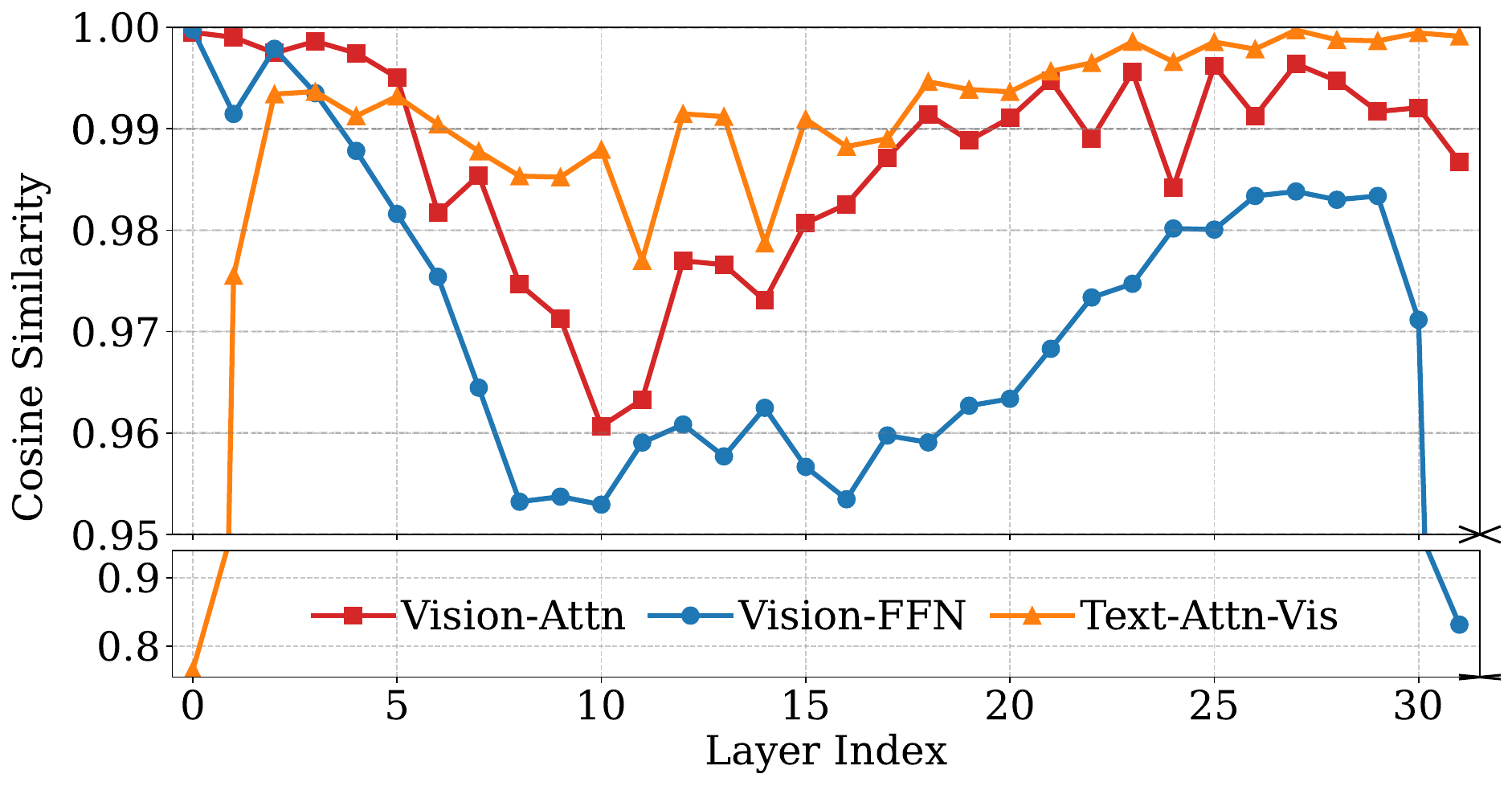}
    \caption{LLaVA-1.5-7B}
\end{subfigure}
\hfill
\begin{subfigure}[t]{0.32\linewidth}
    \centering
    \includegraphics[width=\linewidth]{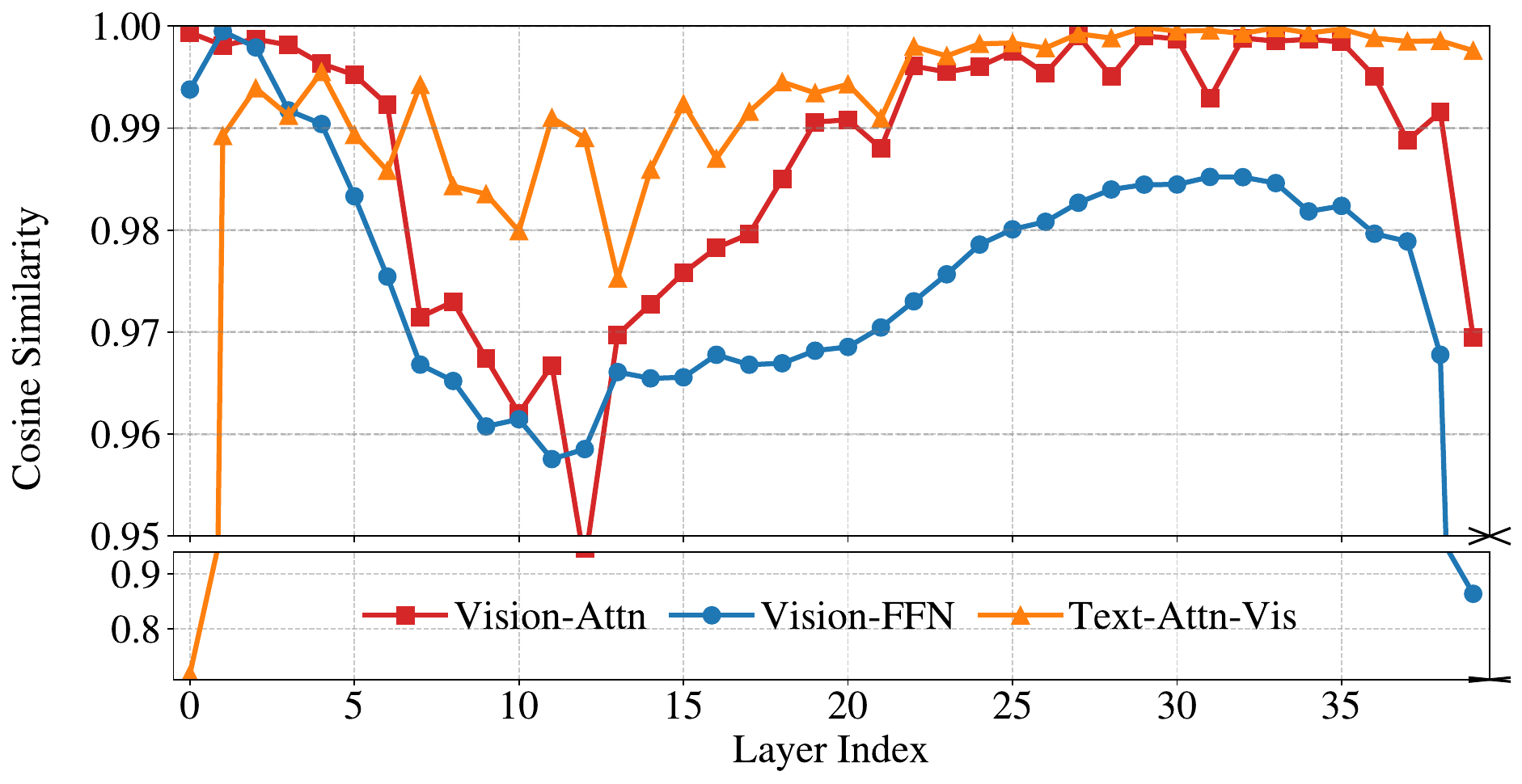}
    \caption{LLaVA-1.5-13B}
\end{subfigure}

\caption{\small
Token-wise mean cosine similarity on TextVQA~\cite{singh2019TextVQA} before and after key modules across LLaVA-1.5-3B, 7B,and 13B. \textit{Vision-Attn} measures vision token similarity before and after attention, \textit{Vision-FFN} measures vision token similarity before and after FFN, and \textit{Text-Attn-Vis} measures text-token similarity before and after text--vision cross-attention.}
\label{fig:cosine similarity}
\end{figure*}

\subsection{Layerwise Redundancy in Text-to-Vision Cross-Attention}
\label{app:t2v-analysis}

We compute the cosine similarity of text-token representations before and after vision-to-text cross-attention at each layer to identify critical text-to-vision cross-attention layers. As shown in Figure~\ref{fig:cosine similarity}, \textit{Text-Attn-Vis} represents the cosine similarity of text tokens before and after \textit{text--vision} cross-attention. The similarity is lowest in the shallow layers (0--1), indicating that visual information is integrated into text representations early. A secondary dip occurs in the middle layers (7--11 and 14), suggesting further text--vision fusion. These shallow and middle layers (0--1, 7--11, and 14) thus serve as the primary stages of text--vision interaction. Beyond these layers, the similarity increases significantly, indicating that the influence of visual information on text representations diminishes.

\begin{table*}[htbp]
\centering
\resizebox{\textwidth}{!}{
\begin{tabular}{lcc|ccccccccc|c}
\toprule
\textbf{Setup} & \textbf{\makecell{Retained Layers}} & \textbf{\makecell{Layer Indices}} &
\textbf{MME$^{P}$} & \textbf{MMB} & \textbf{MMB$^{CN}$} & \textbf{GQA} & \textbf{VQA$^{v2}$} &
\textbf{SQA$^{I}$} & \textbf{VQA\textsuperscript{T}} & \textbf{POPE} & \textbf{SEED$^{I}$} &
\textbf{Avg(\%)} \\
\midrule
\addlinespace[0.4ex] \rowcolor{gray!20}
\multicolumn{13}{c}{\textbf{Baseline}} \\
\midrule
- & Full & 0--31 & 1506.5 & 64.7 & 58.1 & 61.9 & 78.5 & 69.5 & 58.2 & 86.8 & 66.2 & 100.0 \\
\midrule
\addlinespace[0.4ex] \rowcolor{gray!20}
\multicolumn{13}{c}{\textbf{Text--Vision Cross-Attention}} \\
\midrule
\textbf{Diagnosis} &
\multicolumn{12}{c}{
Lowest cosine similarity at layers 0--1, 7--11, and 14 (16 as extension).
} \\
\midrule
\multirow{3}{*}{\textbf{Training-free}} & Essential & 0--1,7--11,14 & 1393.5 & 62.5 & 55.9 & 54.4 & 66.7 & 68.6 & 51.3 & 86.1 & 64.4 & 93.5 \\
& Non-essential & 2--6,12--13,15--31 & 758.2 & 24.9 & 17.4 & 44.8 & - & 61.0 & 45.5 & 70.4 & 43.4 & 63.0 \\
& Frozen & - & 572.1 & 16.0 & 11.3 & 29.9 & 34.0 & 56.5 & 29.9 & 50.3 & 34.2 & 46.2 \\
\midrule
\multirow{2}{*}{\textbf{Training}} & Essential & 0--1, 7--11, 14 & 1476.6 & 66.1 & 58.3 & 62.7 & 77.9 & 67.5 & 56.4 & 87.1 & 66.0 & 99.5 \\
& Essential (expanded) & 0--1, 7--11, 14, 16 & 1463.7 & 69.3 & 59.1 & 62.7 & 78.2 & 68.6 & 56.9 & 87.1 & 66.6 & 100.5 \\
\midrule
\addlinespace[0.4ex] \rowcolor{gray!20}
\multicolumn{13}{c}{\textbf{vision update Attention}} \\
\midrule
\textbf{Diagnosis} &
\multicolumn{12}{c}{
Lowest cosine similarity at layers 6--17. Considering overlap with text--vision fusion, layers 6--11 are identified as key layers.
} \\
\midrule
\multirow{4}{*}{\textbf{Training-free}} & Over-approx. & 6--17 & 1469.7 & 64.7 & 57.1 & 61.6 & 78.1 & 68.5 & 56.6 & 86.0 & 65.6 & 98.8 \\
& Essential & 6--11 & 1454.5 & 64.4 & 57.1 & 61.0 & 77.1 & 68.6 & 56.1 & 86.7 & 65.5 & 98.3 \\
& Non-essential & 0--5,12--31 & 1361.2 & 61.3 & 55.7 & 59.2 & - & 68.8 & 55.4 & 83.9 & 62.9 & 95.3 \\
& Frozen & - & 1298.2 & 62.5 & 54.2 & 54.5 & 70.4 & 69.0 & 51.0 & 79.8 & 61.0 & 91.7 \\
\midrule
\textbf{Training} & Frozen & - & 1501.1 & 66.1 & 59.5 & 62.1 & 78.0 & 69.3 & 57.1 & 87.1 & 65.8 & 100.2 \\
\midrule
\addlinespace[0.4ex] \rowcolor{gray!20}
\multicolumn{13}{c}{\textbf{vision update FFN}} \\
\midrule
\textbf{Diagnosis} &
\multicolumn{12}{c}{
Lowest similarity at layers 5--31. Considering overlap with text--vision fusion, layers 5--11 are identified as key layers.
} \\
\midrule
\multirow{3}{*}{\textbf{Training-free}} & Essential & 5--11 & 1463.2 & 64.5 & 55.2 & 58.9 & 74.6 & 68.7 & 54.2 & 85.2 & 65.3 & 96.8 \\
& Non-essential & 0--4,12--31 & 988.6 & 56.9 & 47.8 & 49.8 & - & 66.3 & 48.9 & 65.4 & 55.2 & 81.8 \\
& Frozen & - & 876.8 & 52.0 & 38.8 & 39.6 & 45.2 & 66.3 & 40.9 & 61.1 & 51.6 & 71.2 \\
\midrule
\multirow{3}{*}{\textbf{Training}} & Essential & 5--11 & 1475.7 & 66.7 & 58.4 & 63.3 & 78.7 & 68.4 & 58.2 & 87.3 & 66.9 & 100.5 \\
& Essential(aggressive) & 5,7,9 & 1493.5 & 67.4 & 57.7 & 62.3 & 78.0 & 70.8 & 57.9 & 87.4 & 65.8 & 100.5 \\
& Frozen & - & 1472.4 & 65.6 & 56.9 & 62.0 & 77.5 & 70.3 & 56.9 & 87.3 & 64.6 & 99.2 \\
\bottomrule
\end{tabular}
}
\caption{\small
Redundancy diagnosis for identifying and validating candidate key layers of vision-related operations in LLaVA-1.5-7B. Step~1 measures cosine similarity between tokens before and after the target operations to identify candidate key layers. In Step~2, we perform training-free verification by retaining only the critical layers and comparing against retaining only the remaining non-critical layers and masking all layers. Step~3 conducts retraining experiments comparing models retaining only the key layers to those removing all layers.
}
\label{tab:llava7b_trainfree}
\end{table*}

We further conduct \textit{training-free masking} experiments, summarized in Table~\ref{tab:llava7b_trainfree}. Masking all text-to-vision cross-attention layers causes a 53.8\% drop in performance across nine benchmarks. In contrast, retaining text-to-vision cross-attention only in the selected shallow and middle layers (0--1, 7--11, and 14) limits the performance drop to 6.5\%, while masking cross-attention in these layers results in a much larger drop of 37.0\%. These results indicate that text-to-vision cross-attention is essential for visual fusion, but its effect is concentrated in the key layers mentioned above, with the remaining layers contributing marginally to final multimodal reasoning.

Building on these results, we validate the findings under \textit{retraining}, as shown in Table~\ref{tab:llava7b_trainfree}. Enabling text-to-vision cross-attention only in the selected shallow and middle layers (0--1, 7--11, and 14) restores 99.5\% of the baseline performance on average. Adding a low-similarity layer (16) further improves performance, slightly surpassing the baseline at 100.5\%. These findings demonstrate that preserving text-to-vision cross-attention only in the essential shallow and middle layers is sufficient to recover the original performance.

\begin{tcolorbox}
\textbf{Finding 1:} Multimodal reasoning depends on only a small number of \textit{text-to-vision cross-attention} layers that inject visual information into text tokens.
\end{tcolorbox}

\subsection{Layerwise Redundancy in Attention-Based Vision Updates}
\label{app:v2v-analysis}

After analyzing text-to-vision cross-attention, we investigate the dynamics of visual representations by computing the cosine similarity of vision token representations before and after the attention module at each layer. As shown in Figure~\ref{fig:cosine similarity}, \textit{Vision-Attn} exhibits lower similarity in the middle layers (6--17), indicating continuous changes in visual representations. However, since text-to-vision cross-attention is mainly concentrated in layers 7--11, vision updates in layers 12--17 are less likely to influence text tokens through cross-modal pathways, thus having a limited impact on final predictions.

To verify this, we perform \textit{training-free masking} experiments, as shown in Table~\ref{tab:llava7b_trainfree}. Masking all \textit{vision update attention} operations results in a 9.3\% performance drop. Retaining updates in layers 6--17 reduces the drop to 1.2\%, while restricting them to layers 6--11 yields a similar drop. In contrast, masking updates in layers 6--11 alone causes a larger drop of 4.7\%. Overall, compared to text-to-vision cross-attention, attention-based updates on vision tokens contribute much less to performance, with layers 6--11 being relatively more important.

Finally, we validate this conclusion by retraining the model. In the training-free setting, masking all attention-based vision updates causes a smaller drop than masking text--vision cross-attention. Based on this observation, and noting that \textit{vision self-attention} accounts for most of the attention computation, we remove all attention operations that update vision tokens and retrain the model. As shown in Table~\ref{tab:llava7b_trainfree}, the retrained model achieves 100.2\% of baseline performance across nine benchmarks, indicating that attention-based vision updates are functionally ineffective. This suggests that vision-related quadratic computation can be safely reduced to linear scale.

\begin{tcolorbox}
\textbf{Finding 2:} Attention-based vision updates in later layers have limited impact on final predictions, and vision-related quadratic computations can be reduced to linear scale without significant performance loss.

\end{tcolorbox}

\subsection{Layerwise Redundancy in FFN-Based Vision Updates}
\label{app:ffn-analysis}
Since FFNs have a large expansion dimension, they are the primary linear component of vision-side computation, with detailed FLOPs derivations provided in Appendix~\ref{sec:Calculation-Equation}. In MLLMs, FFNs primarily refine visual features, while cross-modal information exchange occurs in attention. We hypothesize that FFN updates on vision tokens after the effective text-to-vision cross-attention layers, confined to shallow and middle layers, have limited influence on text representations and final outputs.

To test this hypothesis, we use the same cosine-similarity method to compute the mean similarity of vision token representations before and after the FFN at each layer. As shown in Figure~\ref{fig:cosine similarity}, \textit{Vision-FFN} represents the cosine similarity of vision tokens before and after each FFN layer. The similarity remains low after layer 5, indicating representational changes. Since text-to-vision cross-attention is most active in layers 7 to 11, this suggests that FFN updates in layers 5 to 11 play a significant role in refining visual features. Updates in deeper layers are less likely to influence text tokens and final outputs.

We conduct \textit{training-free pruning} experiments by skipping FFN computation on vision tokens in selected layers. As shown in Table~\ref{tab:llava7b_trainfree}, skipping FFNs across all layers results in a 28.8\% performance drop. In contrast, retaining FFNs only in layers 5 to 11 reduces the drop to 3.2\%. Skipping FFNs in layers 5 to 11 results in an 18.2\% drop. This indicates that only a small subset of FFN layers is necessary for vision token updates.

\begin{table*}[t]
\centering
\resizebox{\textwidth}{!}{
\begin{tabular}{lcc|ccccccccc|c}
\toprule
\textbf{Setup} & \textbf{\makecell{Retained Layers}} & \textbf{\makecell{Layer Indices}} &
\textbf{MME$^{P}$} & \textbf{MMB} & \textbf{MMB$^{CN}$} & \textbf{GQA} & \textbf{VQA$^{v2}$} &
\textbf{SQA$^{I}$} & \textbf{VQA\textsuperscript{T}} & \textbf{POPE} & \textbf{SEED$^{I}$} &
\textbf{Avg(\%)} \\
\midrule
\addlinespace[0.4ex] \rowcolor{gray!20}
\multicolumn{13}{c}{\textbf{Baseline}} \\
\midrule
- & Full & 0--31 & 1529.9 & 68.5 & 63.5 & 63.3 & 80.0 & 72.8 & 61.2 & 87.0 & 68.2 & 100.0 \\
\midrule
\addlinespace[0.4ex] \rowcolor{gray!20}
\multicolumn{13}{c}{\textbf{Text--Vision Cross-Attention}} \\
\midrule
\textbf{Diagnosis} &
\multicolumn{12}{c}{
Lowest similarity at layers 0, 6, 8--10, 13-14, and 16.
} \\
\midrule
\multirow{2}{*}{\textbf{Training-free}} & Essential & 0,6,8--10,13--14,16 & 1362.8 & 66.4 & 61.7 & 56.8 & -- & 71.3 & 55.3 & -- & 66.8 & 94.2  \\
& Frozen & - & 673.3 & 15.4 & -- & 26.7 & 69.5 & 56.5 & 22.9 & 50.6 & 33.1 & 52.1 \\

\midrule
\addlinespace[0.4ex] \rowcolor{gray!20}
\multicolumn{13}{c}{\textbf{vision update Attention}} \\
\midrule
\textbf{Diagnosis} &
\multicolumn{12}{c}{
Lowest similarity at layers 7--18. Considering overlap with text--vision fusion, layers 7--14 are identified as key layers.
} \\
\midrule
\multirow{3}{*}{\textbf{Training-free}} & Over-approx. & 7--18 & 1461.1 & 68.0 & 62.8 & 62.6 & 79.3 & 71.9 & 59.5 & 85.7 & 67.7 & 98.4 \\
& Essential & 7--14 & 1464.4 & 68.1 & 62.6 & 62.2 & 79.0 & 71.9 & 59.5 & 86.1 & 67.6 & 98.3 \\
& Frozen & - & 1246.1 & 63.9 & 55.0 & 54.8 & 69.5 & 69.1 & 47.8 & 87.2 & 63.4 & 89.0 \\
\midrule
\addlinespace[0.4ex] \rowcolor{gray!20}
\multicolumn{13}{c}{\textbf{vision update FFN}} \\
\midrule
\textbf{Diagnosis} &
\multicolumn{12}{c}{
Lowest similarity at layers 5--39. Considering overlap with text--vision fusion, layers 5--14 are identified as key layers.
} \\
\midrule
\multirow{2}{*}{\textbf{Training-free}} & Essential & 5--14 & 1505.0 & 68.8 & 62.6 & 62.6 & 79.1 & 72.6 & 59.5 & 87.2 & 67.8 & 99.1 \\ 
& Frozen & - & 969.5 & 61.3 & 52.9 & 49.2 & 54.3 & 71.3 & 48.9 & 57.6 & 57.9 & 79.0 \\
\bottomrule
\end{tabular}
}
\caption{\small
Redundancy diagnosis for identifying and validating candidate key layers of vision-related operations in LLaVA-1.5-13B.
Step~1 measures cosine similarity using token before and after the target operations to identify candidate key layers. 
Step~2 performs a training-free verification by retaining candidate key layers and comparing with masking all layers. 
}
\label{tab:llava13b_trainfree}
\end{table*}

Finally, we validate these observations by retraining the model. In the retraining process, vision tokens skip FFN computation in selected layers, and we evaluate the model on nine benchmarks. As reported in Table~\ref{tab:llava7b_trainfree}, skipping all FFN layers for vision tokens still preserves 99.2\% of the baseline performance. Enabling FFNs only in layers 5 to 11, or selectively in layers 5, 7, and 9, both reaches 100.5\% of the baseline.

\begin{tcolorbox}
\textbf{Finding 3:} Skipping all FFN updates on vision tokens, even after retraining, still preserves near-baseline performance, indicating that only a small subset of FFN layers is crucial for vision token updates.
\end{tcolorbox}

\subsection{Redundancy Diagnosis Across Model Scales}
\label{app:across-model-analysis}

The findings above are first established on LLaVA-1.5-7B. To examine whether these redundancy patterns generalize across different model scales, we apply the same layerwise redundancy diagnosis to LLaVA-1.5-3B and LLaVA-1.5-13B, assessing whether the pruning strategy developed for 7B can be transferred to these models.

We present the layerwise cosine-similarity curves for LLaVA-1.5-3B and 13B in Figure~\ref{fig:cosine similarity}, respectively, analyzing the same three components:
(i) text-to-vision cross-attention,
(ii) vision update attention, and
(iii) vision update FFN.

For text-to-vision cross-attention, low-similarity layers are observed at \{0--1, 14--15, 18--19, 21--23\} in LLaVA-1.5-3B and at \{0, 6, 8--10, 13--14, 16\} in LLaVA-1.5-13B. Training-free masking experiments (Table~\ref{tab:llava13b_trainfree}, Table~\ref{tab:llava3b_trainfree}) show that retaining only these candidate key layers preserves 87.1\% and 94.2\% of the original average accuracy for 3B and 13B, respectively.

\begin{table*}[htbp]
\centering
\resizebox{\textwidth}{!}{
\begin{tabular}{lcc|cccccccc|c}
\toprule
\textbf{Setup} & \textbf{\makecell{Retained Layers}} & \textbf{\makecell{Layer Indices}} &
\textbf{MME$^{P}$} & \textbf{MMB} & \textbf{MMB$^{CN}$} & \textbf{GQA} & \textbf{SQA$^{I}$} & \textbf{VQA\textsuperscript{T}} & \textbf{POPE} & \textbf{SEED$^{I}$} &
\textbf{Avg(\%)} \\
\midrule
\addlinespace[0.4ex] \rowcolor{gray!20}
\multicolumn{12}{c}{\textbf{Baseline}} \\
\midrule
- & Full & 0--31 & 1258.2 & 57.0 & 28.6 & 59.4 & 60.0 & 48.6 & 86.4 & 60.1 & 100.0 \\
\midrule
\addlinespace[0.4ex] \rowcolor{gray!20}
\multicolumn{12}{c}{\textbf{Text--Vision Cross-Attention}} \\
\midrule
\textbf{Diagnosis} &
\multicolumn{11}{c}{
Lowest similarity at layers 0--1, 14--15, 18--19, and 21--23.
} \\
\midrule
\multirow{2}{*}{\textbf{Training-free}} & Essential & 0--1,14--15,18--19,21--23 & 1156.0 & 50.7 & 18.3 & 49.5 & 58.7 & 41.2 & 80.3 & 57.3 & 87.1  \\
& Frozen & - & 558.2 & 11.0 & 4.0 & 31.9 & 52.9 & 22.3 & 49.7 & 33.4 & 47.3 \\
\midrule
\addlinespace[0.4ex] \rowcolor{gray!20}
\multicolumn{12}{c}{\textbf{vision update Attention}} \\
\midrule
\textbf{Diagnosis} &
\multicolumn{11}{c}{
Lowest similarity at layers 2--3 and 13--31.
Considering overlap with text--vision fusion, layers 2--3 and 13--23 are identified as key layers.
} \\
\midrule
\multirow{3}{*}{\textbf{Training-free}} & Over-approx. & 2--3,13--31 & 1295.9 & 54.6 & 31.5 & 57.6 & 59.8 & 45.9 & 84.7 & 58.7 & 99.5 \\
& Essential & 2--3,13--23 & 1293.8 & 54.7 & 31.1 & 57.6 & 59.8 & 45.8 & 84.6 & 58.9 & 99.3 \\
& Frozen & - & 901.1 & 18.3 & 3.2 & 41.0 & 53.8 & 34.6 & 77.4 & 39.3 & 62.5 \\
\midrule
\addlinespace[0.4ex] \rowcolor{gray!20}
\multicolumn{12}{c}{\textbf{vision update FFN}} \\
\midrule
\textbf{Diagnosis} &
\multicolumn{11}{c}{
Lowest similarity at layers 3--31.
Considering overlap with text--vision fusion, layers 3--23 are identified as key layers.
} \\
\midrule
\multirow{2}{*}{\textbf{Training-free}} & Essential & 3--23 & 1305.4 & 57.0 & 32.9 & 58.4 & 61.5 & 46.9 & 86.3 & 59.5 & 101.9 \\ 
& Frozen & - & 580.3 & 19.3 & 5.2 & 34.6 & 53.7 & 28.6 & 48.5 & 37.1 & 52.8  \\
\bottomrule
\end{tabular}
}
\caption{\small
\textbf{Redundancy diagnosis for identifying and validating key vision-related operations in LLaVA-1.5-3B.}
\textbf{Step~1} measures cosine similarity using token before and after the target operations to identify key layers. 
\textbf{Step~2} performs a training-free verification by masking non-critical layers and comparing with masking all layers. 
}
\label{tab:llava3b_trainfree}
\end{table*}

The cosine-similarity curves of vision update attention exhibit broad low-similarity regions (3B: \{2--3, 13--31\}; 13B: \{7--18\}). When aligned with the effective text-to-vision cross-attention window, these regions become substantially narrower. Retaining vision update attention only in the identified key layers preserves 99.3\% and 98.3\% accuracy for the 3B and 13B models, respectively, while masking all vision update attention leads to degradations of 37.5\% and 11.0\%, which remain smaller than those caused by removing all text-to-vision cross-attention.

Similarly, vision FFNs exhibit broad low-similarity ranges (3B: \{3--31\}; 13B: \{5--39\}), but their effective contribution becomes sharply constrained once aligned with the text-to-vision cross-attention window. Training-free pruning shows that enabling FFNs only in the identified layers preserves 101.9\% and 99.1\% of the average accuracy for the 3B and 13B models, respectively, indicating that only a small subset of FFNs meaningfully contributes to multimodal reasoning. In contrast, skipping all vision FFNs leads to substantially larger performance degradations of 47.2\% and 21.0\%.

Based on these consistent patterns, we transfer the pruning strategy developed for LLaVA-1.5-7B to the 3B and 13B models with standard pretraining and finetuning. As shown in Sec.~\ref{sec:ablation_studies} and Table~\ref{tab:all_ablation}, removing visual token updates in both attention and FFN modules yields the \textit{FreezeVis} variants, which recover 100.3\% and 97.4\% of the baseline accuracy, respectively. Further restricting visual key--value participation to the identified critical text-to-vision cross-attention layers produces the \textit{ViCA} variants, retaining 98.7\% and 97.0\% accuracy while reducing vision-related computation to 4.5\% and 3.3\%. These results demonstrate that the proposed minimal and efficient architecture generalizes consistently across model scales.

\section{Implementation Details}
\label{app:imp}

\subsection{Details of Backbones}
\label{app:details_llm}
Table~\ref{tab:model_config} summarizes the backbone configurations used in our experiments.

\begin{table}[htbp]
\centering
\resizebox{1.0\linewidth}{!}{
\begin{tabular}{lccc}
\toprule
Model & LLaVA-3B & LLaVA-7B & LLaVA-13B \\
\midrule
LLM Backbone & MobileLLaMA-2.7B & Vicuna-7B & Vicuna-13B \\
Blocks      & 32               & 32        & 40          \\
Heads       & 32               & 32        & 40          \\
Hidden Dim  & 2560             & 4096      & 5120        \\
FFN Dim     & 6912             & 11008     & 13824       \\
\bottomrule
\end{tabular}
}
\caption{\small Model configurations of LLaVA-1.5 backbones.}
\label{tab:model_config}
\end{table}

\subsection{Configuration Details of Our Method}
\label{sec:pruning-configurations}

\paragraph{Preserved Text-to-Vision Cross-Attention Layers.}
\label{sec:key_layers}
Preserved text-to-vision cross-attention layers for each backbone:
\begin{itemize}
    \item \textbf{LLaVA-1.5-3B}: \(\{0, 1, 14, 15, 18, 19, 21, 22, 23\}\).
    \item \textbf{LLaVA-1.5-7B}: \(\{0, 1, 7, 8, 9, 10, 11, 14\}\).
    \item \textbf{LLaVA-1.5-13B}: \(\{0, 6, 8, 9, 10, 13, 14, 16\}\).
\end{itemize}

\paragraph{Integration of PyramidDrop into Our Method.}
\label{sec:pdrop}
During inference, our method integrates PyramidDrop, which applies multi-stage vision token dropping for each backbone:
\begin{itemize}
    \item \textbf{LLaVA-1.5-3B}: 25\% of vision tokens are dropped at layers 1, 14, and 18.
    \item \textbf{LLaVA-1.5-7B}: 25\% of vision tokens are dropped at layers 1, 7, and 10.
    \item \textbf{LLaVA-1.5-13B}: 25\% of vision tokens are dropped at layers 6, 9, and 13.
\end{itemize}

\subsection{Details of Calculation Equation of FLOPs}
\label{sec:Calculation-Equation}

To further explore the computation bottleneck, we refer to the theoretical formulations of vision-related computation proposed in prior works \cite{chen2025fastv, yang2024visionzip, liu2025fine, zhang2025VScan}, and decompose the vision-related computation in MLLMs into two major components.
\paragraph{Attention Computation.}
In each attention block, input tokens are first linearly projected into queries, keys, and values, followed by attention multiplications \(QK^\top V\) that enable information exchange among tokens.  
The resulting outputs are then linearly projected by an output projection to produce the final representations of the block.  

Among these operations, the visual-related projections include four components---\(W_Q\), \(W_K\), \(W_V\), and \(W_O\).  
Thus, the total cost of visual attention projections is:
\begin{equation}
\mathcal{C}_{\text{Vis-Projector}} = n_{\text{layers}}* 2* 4 n d^2
\end{equation}
where \(n_{\text{layers}}\) is the number of transformer blocks,
the factor of 2 converts MACs into FLOPs,
and \(n\) denotes the number of vision tokens while \(d\) denotes the hidden dimension.

During the attention multiplication stage, vision tokens engage in two types of interactions:  
(i) \textit{vision-to-vision self-attention},  and  
(ii) \textit{text-vision cross-attention},  
with corresponding computational costs:
\begin{equation}
\begin{aligned}
\mathcal{C}_{\text{vision-vision}} &= n_{\text{layers}}* 2* 2dn^2 \\
\mathcal{C}_{\text{text-vision}} &= n_{\text{layers}}* 2* 2dnt
\end{aligned}
\end{equation}
where \(t\) denote the numbers of text tokens.  
The total attention multiplication cost about vision tokens is:
\begin{equation}
\mathcal{C}_{Vis-QK^{\top}V} = n_{\text{layers}}* 2* 2d(n^2  + nt)
\end{equation}
Accordingly, the total visual-related attention cost can be expressed as:
\begin{align}
\mathcal{C}_{\text{Vis-Attn}} 
&= \mathcal{C}_{\text{Vis-Projector}} + \mathcal{C}_{\text{Vis-}QK^{\top}V} \nonumber\\
&= n_{\text{layers}}* 2*( 4nd^2 + (2dn^2  + 2dnt))
\end{align}

\paragraph{FFN Computation.}
Each Feed-Forward Network (FFN) comprises three main modules:  
(i) an \textit{up-projection} that expands the hidden dimension from \(d\) to \(m\),  
(ii) a \textit{gating} operation, and  
(iii) a \textit{down-projection} that maps features back from \(m\) to \(d\).  
Hence, the total cost of a gated FFN is:
\begin{equation}
\mathcal{C}_{\text{Vis-FFN}} = n_{\text{layers}}* 2*( 3n d m)
\end{equation}
where \(m\) denote the expansion dimension.  
Since \(m \gg d\), FFNs constitute the dominant linear component of the overall visual computation.

\paragraph{Total Computation.}
By combining the attention and FFN costs, the total visual computation can be summarized as:
\begin{align}
\mathcal{C}_{\text{Vis}}
&= \mathcal{C}_{\text{Vis-Attn}} + \mathcal{C}_{\text{Vis-FFN}} \nonumber \\
&= n_{\text{layers}}* 2*( 4nd^2 + 2d(n^2 + nt) + 3ndm)
\label{eq:Vis-Total}
\end{align}
This formulation clearly shows that the vision self-attention term grows quadratically with the number of vision tokens, while all other terms scale only linearly with \(n\).  
When \(n \ll d\), the quadratic term \(2dn^2\) is substantially smaller than the attention projection cost \(4nd^2\) and the FFN cost \(3ndm\), making the total visual computation scale \textit{near linearly} with the number of vision tokens. Extending this analysis, the total computation is:
\begin{equation}
\mathcal{C}_{\text{Total}}
= n_{\text{layers}} \cdot 2\big(4Ld^2 + 2dL^2 + 3Ldm\big)
\label{eq:Total-FLOPs}
\end{equation}
where \(L = n  + t\).  
Here, \(4Ld^2\) corresponds to the four projection operations
(\(W_Q\), \(W_K\), \(W_V\), \(W_O\)) applied to all tokens,  
\(2dL^2\) represents the FLOPs of the \(QK^{\top}V\) attention
multiplications across all tokens,  
and \(3Ldm\) denotes the computation of the FFN module for the entire sequence. 
To further quantify the contribution of visual computation, we examine the ratio between the visual computation and the total FLOPs:
\begin{equation}
\frac{\mathcal{C}_{\text{Vis}}}{\mathcal{C}_{\text{Total}}}
=
\frac{4nd^2 + 2d(n^2  + nt) + 3ndm}
     {4Ld^2 + 2dL^2 + 3Ldm}
\label{eq:Vis-Ratio}
\end{equation}

\paragraph{Equivalent Visual Token Count.}
\label{para:equiv-token-count}
To enable fair comparison, we report the \textit{equivalent visual token count} \(n_{\text{eq}}\), which is the number of visual tokens that would produce the same vision-side FLOPs in the original model. Let \(N=n_{\mathrm{layers}}\) and
\begin{equation}
A = 4d^2 + 2dt + 3dm .
\label{eq:eq-token-a}
\end{equation}
Then \(n_{\text{eq}}\) is computed as:
\begin{equation}
D = \sqrt{(2NA)^2 + 16Nd\,\mathcal{C}_{\mathrm{Vis}}}
\label{eq:eq-token-d}
\end{equation}
\begin{equation}
n_{\text{eq}}
=
\frac{D - 2NA}{8Nd}
\label{eq:eq-token-count}
\end{equation}

\begin{figure}[t]
\centering
\includegraphics[width=1.0\linewidth]{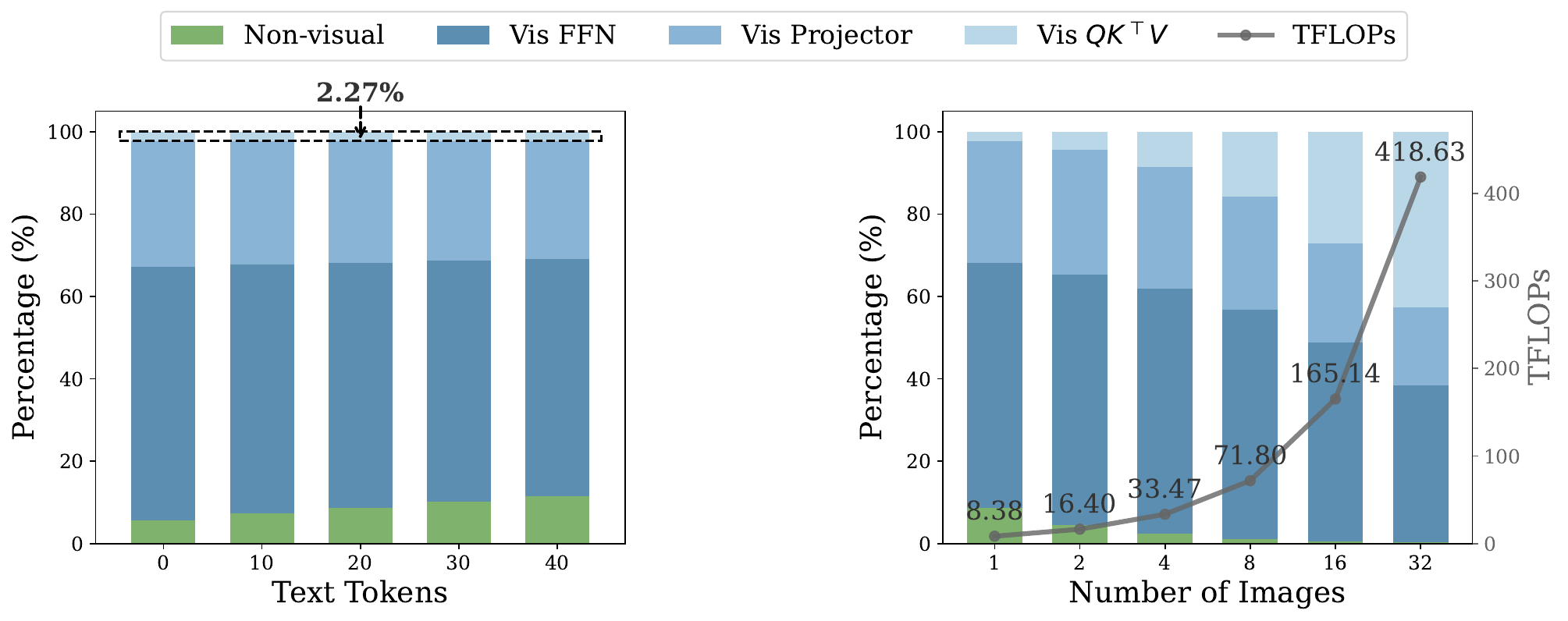}
\caption{
\textbf{Computation breakdown of LLaVA-1.5-7B across question and vision token scales.}
\textit{Vis $QK^{\top}V$}, \textit{Vis Projector}, \textit{Vis FFN}, \textit{Non-Visual}, and \textit{TFLOPs} denote the costs of visual-related attention multiplications, visual attention projections, visual FFN, non-visual computation, and total computation, respectively.
Left: \textit{Vis $QK^{\top}V$} stays at 2.27\%.
}
\label{fig:Decomposition-Vision-Computation}
\end{figure}

\paragraph{Case Study on LLaVA-1.5-7B.} We further analyze the computational cost in LLaVA-1.5-7B, 
using representative parameters \(n = 576\), \(d = 4096\), \(m = 11008\), and \(n_{{layers}} = 32\). 
The number of text tokens $  t  $ consists of system tokens $  t_s = 35  $ and question tokens $  t_t  $, where $  t_t  $ typically ranges from 0 to 40 across benchmarks. 
Among all vision-related components, only the text-to-vision cross-attention in \textit{\(QK^{\top}V\)} attention multiplication depends on the $  t_t  $. 
As illustrated in Figure~\ref{fig:Decomposition-Vision-Computation}, as \(t_t\) increases from 0 to 40, 
the proportion of \textit{Vis \(QK^{\top}V\)} multiplication in the total computation
consistently remains around 2.27\%, indicating that variations in question tokens length
(from 0 to 40 tokens) have a negligible effect. 
Therefore, for simplicity, we fix \(t_t = 20\) in all subsequent analyses.

In contrast to the negligible effect of text length, the number of vision tokens strongly reshapes the computational profile.
As shown in Figure~\ref{fig:Decomposition-Vision-Computation}, increasing the image count from 1 to 32 (576 tokens each) shifts the bottleneck from FFN to attention:
the visual FFN share drops from 59.49\% to 38.12\%, visual attention-projector from 29.52\% to 18.91\%,
while visual $QK^\top V$ multiplications soar from 2.27\% to 42.68\%.
This marks a clear transition from FFN-dominated to attention-dominated computation as visual token density increases.

Using Eq.~(\ref{eq:Vis-Total}) and ~(\ref{eq:Total-FLOPs}), the visual FLOPs of LLaVA-1.5-7B is:
\begin{equation}
\begin{aligned}
\mathcal{C}_{\text{Vis}}
&= n_{\text{layers}}\cdot 2
\big(4nd^2 + 2d(n^2 + nt) + 3ndm\big) \\
&\approx 7.65\ \text{TFLOPs}
\end{aligned}
\end{equation}
and the total FLOPs is:
\begin{equation}
\begin{aligned}
\mathcal{C}_{\text{Total}}
&= n_{\text{layers}}\cdot 2
\big(4Ld^2 + 2dL^2 + 3Ldm\big) \\
&\approx 8.38\ \text{TFLOPs}
\end{aligned}
\end{equation}
where \(L=n+t\), \(t=t_s+t_t\) and we use \(n{=}576\), \(d{=}4096\), \(m{=}11008\), \(s{=}35\), \(t{=}20\), and \(n_{layers}{=}32\).

\paragraph{Cost of visual updating in LLaVA-1.5.}
\label{app:visual-update-ratio}
In LLaVA-1.5, the LLM's update to visual tokens consists of two main parts: self-attention and FFN. The FLOPs for this visual update are expressed as:
\begin{equation}
\begin{aligned}
\mathcal{C}_{\text{Vis-update}}
&= n_{\text{layers}} \times 2 \\
&\quad \times (2nd^2 + 2dn^2 + 3ndm)
\end{aligned}
\end{equation}
The ratio of the visual update computation to the total visual path computation is given by:
\begin{equation}
\text{Ratio}
=
\frac{2nd^{2} + 2dn^{2} + 3ndm}
{4nd^{2} + 2d(n^{2} + nt) + 3ndm}
\end{equation}
The visual update computation accounts for more than 80\% of the total visual computation. Detailed numerical results are shown in the table~\ref{tab:visual-update-ratio}:

\begin{table}[ht]
\centering
\setlength{\tabcolsep}{3pt}
\small
\begin{tabular}{lccc}
\toprule
Model & \makecell{Visual Update\\TFLOPs} & \makecell{Total Visual\\TFLOPs} & \makecell{Ratio\\(\%)} \\
\midrule
LLaVA-1.5-3B  & 2.55 & 3.04 & 84.0\% \\
LLaVA-1.5-7B  & 6.41 & 7.65 & 83.8\% \\
LLaVA-1.5-13B & 12.48 & 14.91 & 83.7\% \\
\bottomrule
\end{tabular}
\caption{\small Ratio of visual update computation to total visual path computation in LLaVA-1.5.}
\label{tab:visual-update-ratio}
\end{table}

\paragraph{Computation of Our Method.}
After removing redundant visual pathways, only a small number of text--vision cross-attention layers are retained. 
In these layers, the remaining vision-related computation consists solely of:  
(i) the \textit{key--value projection} of vision tokens in attention blocks, and  
(ii) the \textit{text--vision cross-attention} multiplication in attention block.  
Their computational costs become:
\begin{equation}
\begin{aligned}
\mathcal{C}_{\text{Vis-Projector}}
&= n_{\text{retained}}\cdot 2 \cdot 2nd^2 \\
\mathcal{C}_{\text{text-to-vision}}
&= n_{\text{retained}}\cdot 2 \cdot 2dnt_t
\end{aligned}
\end{equation}
where \(n_{\text{retained}}\) denotes the number of text--vision cross-attention layers that are retained.
Using representative LLaVA-1.5-7B parameters (\(n = 576\), \(d = 4096\), \(t_t = 20\)),  
the ratio of these two remaining components is:
\begin{equation}
\frac{\mathcal{C}_{\text{Vis-Projector}}}{\mathcal{C}_{\text{text-to-vision}}}
= \frac{2 n d^2}{2 d n t_t}
\approx 204
\end{equation}
With the above two components being the only remaining vision-related
operations, the overall visual FLOPs after applying our method become:
\begin{equation}
\begin{aligned}
\mathcal{C}_{\text{Vis-Ours}}
&= n_{\text{retained}}\big(2nd^2 + 2dnt_t\big) \\
&\approx 0.31\ \text{TFLOPs}
\end{aligned}
\end{equation}
and the corresponding total FLOPs is:
\begin{equation}
B = 4td^2 + 2dt^{2} + 3tdm
\end{equation}
\begin{equation}
\begin{aligned}
\mathcal{C}_{\text{Total-Ours}}
&= \mathcal{C}_{\text{Vis-Ours}} + 2n_{\text{layers}}B \\
&\approx 1.02\ \text{TFLOPs}
\end{aligned}
\end{equation}
where \(t=s+t\), \(m = 11008\), \(s = 35\), 
\(n_{\text{retained}} = 8\), and \(n_{\text{layers}} = 32\).
At this point, the dominant part of the total computation is contributed by
text inference rather than any vision-related operations in the our pruned model.

\subsection{FlashAttention Acceleration Implementation}
\label{app:flash-atten---imp}

\begin{figure}
    \centering
    \includegraphics[width=1.0\linewidth]{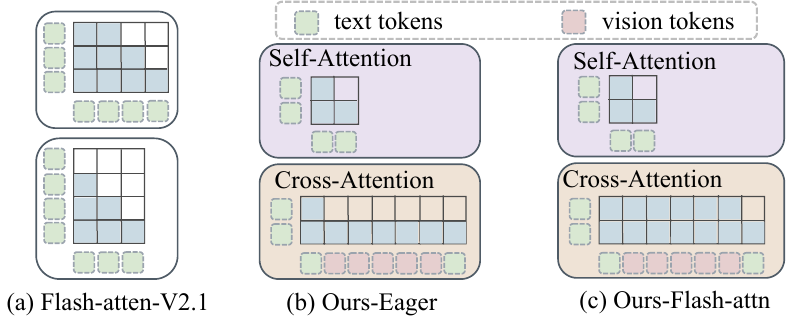}
    \caption{ \small
    (a) Standard FlashAttention causal masking when the query length differs from the key--value length, where the causal mask is aligned to the bottom-right corner of the attention matrix.
    (b) Our method under eager attention, where visual tokens are frozen via explicit masking and participate in attention only as key--value representations.
    (c) Our method under FlashAttention, where visual tokens participate only as key--value and the causal mask is automatically aligned to the bottom-right corner of the attention matrix.
    }
    \label{fig:flash-atten}
\end{figure}

All previous experiments are conducted under the \emph{eager} attention mode, where redundant visual token update paths are removed via explicit masking.
As shown in Figure~\ref{fig:flash-atten}(b), in layers that preserve visual information, visual tokens participate only as key--value pairs in cross-attention, while queries are generated exclusively from text tokens.
As a result, the query length is significantly shorter than the key--value length.

FlashAttention naturally supports such asymmetric attention patterns.
As illustrated in Figure~\ref{fig:flash-atten}(a), since FlashAttention v2.1, when $\text{seqlen}_q \neq \text{seqlen}_k$ and \texttt{causal=True}, the causal mask is aligned to the bottom-right corner of the attention matrix, enabling efficient computation for short-query and long-key--value scenarios.
The attention weights produced by our method under FlashAttention are shown in Figure~\ref{fig:flash-atten}(c).
Compared to eager attention, system-prompt tokens exhibit slightly increased attention to visual tokens.
However, across multiple benchmarks, this difference results in only minor accuracy degradation in practice.
When CUDA utilization is saturated, our approach achieves approximately a $3.5\times$ speedup in the prefill stage, demonstrating the practical efficiency of the proposed architecture under FlashAttention.

\subsection{Training Dataset}
\label{app:training-data}
We follow the default two-stage training data of LLaVA-1.5~\cite{liu2024improved}. In the first stage, we use the LLaVA-Pretrain dataset, i.e., the BLIP-LAION-CC-SBU 558K image-text data, to learn the multimodal projector. This stage aims to align the frozen visual representations with the language model embedding space before visual instruction tuning.

In the second stage, we use the LLaVA-1.5 665K visual instruction tuning mixture to perform multimodal instruction tuning. This dataset contains diverse image-based instruction-following samples from multiple visual-language tasks, enabling the model to acquire general multimodal conversation and reasoning abilities. For a fair comparison, ViCA uses exactly the same training data as the LLaVA-1.5 7B baseline in both stages. The only difference lies in the attention-update strategy: ViCA freezes the update of visual tokens and enables text-to-vision cross-attention only at the selected layers.

\subsection{Training Configuration}
\label{app:training-config}
ViCA follows the two-stage training recipe of LLaVA-1.5~\cite{liu2024improved}, while applying the ViCA attention configuration throughout training:

\paragraph{Stage 1: projector pretraining.}
Different from the original LLaVA-1.5 pretraining, ViCA keeps visual tokens fixed during LLM propagation and enables text-to-vision cross-attention only at the selected layers identified by our method. This allows the projector to adapt to the same visual-token propagation strategy used in subsequent fine-tuning. 

All other settings follow the official LLaVA-1.5 pretraining script, including the LLaVA-Pretrain BLIP-LAION-CC-SBU 558K data, the two-layer GELU MLP projector, one-epoch training, a learning rate of $1\times10^{-3}$, cosine scheduling, a warmup ratio of 0.03, bf16 mixed precision, TF32 acceleration, and DeepSpeed ZeRO-2.

\paragraph{Stage 2: visual instruction fine-tuning.}
In the second stage, we initialize the model with the projector obtained from Stage 1 and fine-tune it on the LLaVA-1.5 665K visual instruction tuning mixture. We follow the official LLaVA-1.5 instruction-tuning recipe, using the same vision tower, projector architecture, image preprocessing, conversation template, and training hyperparameters. The only differences are that ViCA keeps visual tokens fixed during LLM propagation and enables text-to-vision cross-attention only at the selected layers. 

Across both stages, ViCA uses the same training data and hyperparameter settings as LLaVA-1.5. The training protocol differs only in the fixed visual-token propagation strategy, the layer-specific text-to-vision cross-attention design, and the use of a projector pretrained under the ViCA attention configuration.

\section{Details of Comparison Methods}
\label{app:detail_method}

\paragraph{PyramidDrop~\cite{xing2024pyramiddrop}.}
PyramidDrop is an token pruning method that progressively removes less important tokens across transformer layers to reduce computation while preserving performance. In this work, we apply PDrop to LLaVA-1.5 by discarding half of the remaining tokens at the 1/4, 2/4, and 3/4 depths of the model, without introducing additional learnable parameters. This progressive dropping scheme is equivalent to using approximately 270 tokens per layer on average across the entire network.

\paragraph{YOPO~\cite{zhang2024treat}.}
YOPO applies operation-level pruning by reducing visual attention heads,
restricting the local visual attention radius, dropping the last vision-related
layers, and retaining a fixed ratio of FFN neurons. The detailed pruning
settings for each LLaVA-1.5 backbone are summarized in
Table~\ref{tab:llava_summary_yopo}.

\begin{table}[h]
\centering
\footnotesize
\resizebox{1.0\linewidth}{!}{
\begin{tabular}{lccccc}
\toprule
\textbf{Model} & \textbf{Vis-TFLOPs} & \textbf{Heads} & \textbf{Radius} & \textbf{Layers} & \textbf{Neurons} \\
\midrule
LLaVA-1.5-3B    & 0.36 & 16 & 5 & 16 & 25\% \\
LLaVA-1.5-7B    & 0.92 & 16 & 5 & 16 & 25\% \\
LLaVA-1.5-13B   & 1.79 & 20 & 5 & 20 & 25\% \\
\bottomrule
\end{tabular}}
\caption{ \small 
YOPO pruning configurations for LLaVA-1.5 models, including the number of
retained attention heads (\#Heads), local visual attention radius (\#Radius),
dropped vision layers (\#Layers), and retained FFN neuron ratio (\#Neurons).
}
\label{tab:llava_summary_yopo}
\end{table}

\paragraph{ToMe~\cite{bolya2022tome}.}
ToMe merges similar visual tokens inside Vision Transformers via lightweight token matching,
achieving inference acceleration without additional training by reducing redundant token representations.

\paragraph{FastV~\cite{chen2024image}.}
FastV is the first work to identify redundant visual attention in MLLMs. It performs training-free
early-stage token pruning by removing visual tokens with the lowest visual--text attention scores
after shallow layers.

\paragraph{SparseVLM~\cite{zhang2024sparsevlm}.}
SparseVLM observes that instruction tokens contribute unequally to visual token pruning.
It selects instruction tokens most relevant to the visual input and uses their cross-modal attention
to guide adaptive, progressive token pruning.

\paragraph{HiRED~\cite{arif2025hired}.}
HiRED allocates token budgets across spatial image partitions based on CLS attention.
Within each partition, the most informative visual tokens are retained, enabling spatially aware
token reduction for high-resolution inputs.

\paragraph{HiPrune~\cite{liu2025hiprune}.}
HiPrune leverages hierarchical attention patterns in visual encoders and categorizes tokens into
anchor, buffer, and register types. It performs training-free pruning while preserving object-centric
and global contextual information.

\paragraph{FlowCut~\cite{tong2025flowcut}.}
FlowCut formulates token pruning as an information flow preservation problem, aiming to remove
tokens with minimal contribution to the overall information propagation in vision-language models.

\paragraph{HoloV~\cite{zou2025holov}.}
HoloV addresses semantic collapse under aggressive pruning by holistically allocating token budgets
across spatial regions, ensuring global context coverage and reducing redundancy.

\paragraph{VISA~\cite{jiang2025visa}.}
VISA introduces group-wise visual token selection and aggregation via graph summarization.
Instead of simply discarding tokens, VISA aggregates information from removed tokens into retained
tokens, achieving a favorable efficiency--performance trade-off.

\paragraph{D$^2$Pruner~\cite{zhang2025d2pruner}.}
D$^2$Pruner combines debiased importance estimation with structural diversity.
By constructing a hybrid token graph and applying Maximal Independent Set selection,
it jointly preserves semantic importance and spatial diversity.

\paragraph{VScan~\cite{zhang2025VScan}.}
VScan dynamically scans and selects visual tokens according to task relevance,
adapting token retention patterns across different inference stages.

\paragraph{FiCoCo~\cite{han2024filter}.}
FiCoCo proposes a unified training-free paradigm termed \emph{Filter--Correlate--Compress}
for multimodal token reduction, systematically removing redundancy while preserving key
cross-modal correlations.

\paragraph{VisPruner~\cite{zhang2025vispruner}.}
VisPruner performs visual-only token pruning by combining visual attention-based importance
selection with similarity-based duplicate removal, emphasizing visual diversity preservation.

\paragraph{DivPrune~\cite{alvar2025divprune}.}
DivPrune formulates token pruning as a Max-Min Diversity Problem, selecting a subset of visual tokens
by maximizing the minimum pairwise distance among retained tokens.

\paragraph{DOP~\cite{liu2025fine}.}
Depth-wise Operation Pruning (DOP) reduces computation by pruning token-processing operations
rather than tokens alone, reallocating saved computation to more informative token groups.

\paragraph{CDPruner~\cite{zhang2025beyond}.}
CDPruner introduces conditional diversity maximization for token pruning.
It defines visual token similarity conditioned on the instruction context and formulates
token selection as a Determinantal Point Process optimization.

\paragraph{Dynamic-LLaVA~\cite{huang2024dynamic}.}
Dynamic-LLaVA applies dynamic sparsification strategies across both prefill and decoding stages,
adapting token retention based on KV-cache usage to reduce decoding-time overhead.

\paragraph{TwigVLM~\cite{shao2025growing}.}
TwigVLM integrates token pruning with efficient decoding.
It combines twig-guided token pruning with self-speculative decoding to improve both inference
efficiency and generation speed.

\paragraph{VisionZip~\cite{yang2024visionzip}.}
VisionZip exploits the high concentration of visual attention in vision encoders.
It selects dominant tokens and clusters remaining ones to form contextual tokens, preserving
visual information under aggressive compression.

\paragraph{DART~\cite{wen2025stop}.}
DART emphasizes token diversity over individual importance.
It iteratively selects visual tokens that are maximally dissimilar to previously selected tokens,
reducing redundancy and improving coverage.

\paragraph{TokenPacker~\cite{li2025tokenpacker}.}
TokenPacker compresses visual representations at the projector stage using a coarse-to-fine
token formation strategy, enabling compact yet informative visual token representations.

\paragraph{Delta-LLaVA~\cite{zamini2025delta}.}
Delta-LLaVA introduces low-rank delta projections to compress visual tokens into a compact
subspace, maintaining strong performance under fixed small token budgets.

\paragraph{LLaVA-PruMerge~\cite{shang2025llava-prumerge}.}
LLaVA-PruMerge combines pruning and token merging by first removing low-importance tokens
using visual attention and then clustering retained tokens based on similarity.

\paragraph{TRIM~\cite{song2025less}.}
TRIM leverages CLIP-based image--text similarity to assess visual token importance,
pruning tokens that are weakly aligned with user instructions to improve instruction-aware efficiency.

\section{Details of Benchmarks}
\label{app:benchmark}

\paragraph{MME\textsuperscript{P}~\cite{fu2023mme}.}
MME is a widely used benchmark for evaluating multimodal large language models, covering both perceptual and cognitive aspects. In this paper, we consider the perception subset (MME\textsuperscript{P}), which concentrates on visual understanding through a collection of perception-related tasks (commonly grouped into 14 categories).
These tasks include coarse-grained perception such as object existence, counting, spatial position, and color recognition, as well as fine-grained recognition of posters, celebrities, scenes, landmarks, and artworks, together with OCR-related perception. According to the official evaluation protocol, all questions are posed in a binary (yes/no) format, offering a consistent and controlled assessment of visual perceptual capability.

\paragraph{MMB~\cite{liu2024mmbench}.}
MMBench is a hierarchical benchmark developed to evaluate multimodal vision--language models across a broad spectrum of perception and reasoning abilities. Its design organizes capabilities into multiple levels of granularity, allowing performance to be analyzed from coarse competencies to more detailed sub-skills. The benchmark adopts a multiple-choice formulation and applies circular evaluation to mitigate scoring variance, supporting reliable and fair model comparison across diverse multimodal scenarios.

\paragraph{MMB\textsuperscript{CN}~\cite{liu2024mmbench}.}
MMBench-CN serves as the Chinese-language extension of MMBench, aiming to evaluate multimodal understanding under Chinese linguistic settings. It retains the same hierarchical ability structure as the original benchmark while placing emphasis on Chinese-language vision--language alignment. When considered alongside benchmarks in other languages, MMBench-CN also provides insights into multilingual robustness and cross-lingual generalization.

\paragraph{GQA~\cite{hudson2019gqa}.}
GQA is a visual question answering benchmark intended to test structured visual understanding and compositional reasoning. It is built on images from Visual Genome and leverages detailed scene graph annotations that explicitly encode objects, attributes, and relationships. Questions are generated by following well-defined semantic paths over these scene graphs, making GQA particularly suitable for evaluating reasoning grounded in complex visual structures.

\paragraph{VQA\textsuperscript{v2}~\cite{goyal2017VQAv2}.}
VQA-v2 is a large-scale, open-ended visual question answering benchmark that examines how models combine visual perception with language understanding and commonsense knowledge. The dataset consists of real-world images paired with open-ended questions, each annotated with multiple human-provided answers. Its adversarially balanced question design reduces language bias, encouraging models to rely more heavily on visual evidence.

\paragraph{SQA\textsuperscript{I}~\cite{lu2022ScienceQA}.}
ScienceQA is a multimodal question answering benchmark covering topics from natural sciences, language sciences, and social sciences, with questions organized by topics, categories, and skills. In this work, we evaluate the image-based subset (SQA\textsuperscript{I}), which includes only questions accompanied by visual inputs.
This subset emphasizes multimodal comprehension and scientific reasoning, requiring models to integrate visual information with relevant domain knowledge.

\paragraph{TextVQA~\cite{singh2019TextVQA}.}
TextVQA evaluates a model's ability to recognize and reason about text appearing within natural images.
The images contain diverse textual elements, including signs, storefronts, billboards, and product packaging.
Successfully answering the questions often involves combining OCR outputs with visual context and semantic reasoning, making TextVQA a representative benchmark for text-centric multimodal understanding.

\paragraph{POPE~\cite{li-etal-2023-evaluating}.}
POPE is a diagnostic benchmark designed to analyze object hallucination in vision--language models.
It frames hallucination detection as a set of binary questions that ask whether specific objects are present or absent in a given image. By employing multiple sampling strategies and reporting metrics such as accuracy, precision, recall, and F1 score, POPE provides a fine-grained measure of a model's tendency to hallucinate nonexistent objects.

\paragraph{SEED\textsuperscript{I}~\cite{li2024seed}.}
SEEDBench is a multiple-choice benchmark for evaluating multimodal understanding.
In this study, we focus on the image-only subset (SEED\textsuperscript{I}), which targets spatial visual understanding and reasoning. The evaluation covers a range of aspects, including scene understanding, object attributes, spatial location, counting, spatial relations, interactions, visual reasoning, and text recognition, offering a structured assessment of image-based vision--language performance.

\paragraph{OCRBench~\cite{liu2024ocrbench}.}
OCRBench is a comprehensive benchmark designed to evaluate OCR-related capabilities in multimodal large language models. It covers a wide range of text-centric visual understanding tasks, including text recognition, scene text VQA, document question answering, key information extraction, and handwritten mathematical expression recognition. Since many questions require accurate perception of fine-grained textual details, OCRBench provides a direct test of whether vision-side compression preserves text-sensitive visual information.

\paragraph{BLINK~\cite{fu2024blink}.}
BLINK is a perception-oriented benchmark for evaluating fine-grained visual understanding in multimodal large language models. It reformulates classic computer vision tasks into multiple-choice questions, covering abilities such as relative depth estimation, visual correspondence, visual similarity, forensics detection, and multi-view reasoning. These tasks are difficult to solve through language priors alone, making BLINK suitable for assessing the perceptual robustness of compressed visual representations.

\paragraph{V\textsuperscript{*}~\cite{wu2024v}.}
V\textsuperscript{*}Bench evaluates the ability of multimodal models to perform guided visual search over high-resolution and visually crowded images. The benchmark emphasizes locating task-relevant visual evidence and focusing on fine-grained regions needed to answer a question. It is therefore useful for examining whether visual token pruning removes important local details required for precise visual search and detailed perception.

\paragraph{AI2D~\cite{kembhavi2016diagram}.}
AI2D is a diagram understanding benchmark built from illustrative science diagrams and associated question-answer pairs. It requires models to interpret diagram structures, textual labels, object relationships, and symbolic visual elements. Compared with natural-image VQA, AI2D places stronger emphasis on structured visual reasoning and layout comprehension, making it useful for evaluating whether compressed visual tokens preserve diagram-level semantics.

\paragraph{RealWorldQA.~\cite{xai2024grok15v}}
RealWorldQA is a benchmark introduced to evaluate real-world spatial understanding in multimodal models. It consists of real-world images paired with questions that require models to understand everyday scenes, object relations, and spatial configurations. Since it focuses on practical visual understanding in real environments, RealWorldQA provides a complementary test of whether visual compression preserves contextual and spatial cues.

\paragraph{MMStar~\cite{chen2024we}.}
MMStar is a carefully curated multimodal benchmark designed to evaluate vision-indispensable reasoning in large vision--language models. Its samples are selected to reduce language-only shortcuts and data leakage, ensuring that correct answers require genuine visual understanding. The benchmark covers multiple core capabilities and fine-grained evaluation axes, making it suitable for assessing whether vision-side compression weakens the extraction of necessary visual evidence.

\paragraph{InfoVQA~\cite{mathew2022infographicvqa}.}
InfoVQA evaluates visual question answering over infographic images that combine textual, graphical, numerical, and layout-based information. Answering these questions often requires OCR, chart or icon interpretation, layout understanding, and semantic reasoning over visually dense content. Therefore, InfoVQA provides a challenging testbed for examining whether compressed visual representations retain fine-grained information in text-rich and structure-heavy images.

\paragraph{Flickr30k-CIDEr~\cite{young2014image,vedantam2015cider}.}
Flickr30k is an image captioning dataset consisting of natural images paired with human-written descriptions. In this work, we evaluate captioning quality using CIDEr, which measures the consensus between generated captions and human references. This evaluation reflects whether compressed visual tokens retain sufficient object, attribute, and scene-level information for generating detailed and semantically faithful captions.

\paragraph{RefCOCO~\cite{kazemzadeh2014referitgame,yu2016modeling}.}
RefCOCO is a referring expression comprehension benchmark based on natural images, where the model is required to localize the target object described by a natural-language expression. The expressions often involve object categories, attributes, and spatial relationships, requiring accurate alignment between language and image regions. RefCOCO is therefore useful for evaluating whether visual compression preserves the localization cues needed for spatial grounding.

\paragraph{RefCOCO+~\cite{kazemzadeh2014referitgame,yu2016modeling}.}
RefCOCO+ extends the referring expression comprehension setting with expressions that rely less on absolute location words and more on appearance-based descriptions. This makes the benchmark more dependent on fine-grained visual attributes such as color, texture, category, and object-specific appearance. As a result, RefCOCO+ provides a stricter test of whether pruned visual tokens retain detailed object-level representations.

\paragraph{RefCOCOg~\cite{mao2016generation,yu2016modeling}.}
RefCOCOg is a referring expression benchmark characterized by longer and more descriptive expressions than RefCOCO and RefCOCO+. These expressions often contain richer object attributes, relationships, and contextual descriptions, requiring more comprehensive language-guided visual grounding. Evaluating on RefCOCOg helps examine whether vision-side compression affects localization under complex and fine-grained referring expressions.